\documentclass[twoside,11pt]{article}
\usepackage{jmlr2e}
\pdfoutput=1
\usepackage[colorinlistoftodos]{todonotes}
\usepackage{amsmath}
\usepackage{subcaption}
\usepackage[T1]{fontenc} 

\DeclareMathOperator*{\argmax}{arg\,max}

\newcommand*{\bigCI}{%
  \mathrel{\text{%
    {\rotatebox[origin=c]{90}{\resizebox{2.25ex}{1.65ex}{$\vDash$}}}%
  }}%
}

\ShortHeadings{BayesDB: A system for querying the probable implications of data}{Mansinghka et al.}
\firstpageno{1}

\begin{document}

\pagebreak

\title{BayesDB: A probabilistic programming system for querying the
  probable implications of data}

\author{\name Vikash Mansinghka \email vkm@mit.edu \\
\name Richard Tibbetts \email tibbetts@mit.edu \\
\name Jay Baxter \email jbaxter@mit.edu \\
       \addr Computer Science \& Artificial Intelligence Laboratory\\
       Department of Brain \& Cognitive Sciences \\
       Massachusetts Institute of Technology \\
       Cambridge, MA 02139, USA
       \AND
\name Pat Shafto \email p.shafto@louisville.edu \\
\name Baxter Eaves \email b0eave01@louisville.edu \\
        \addr Department of Psychology \\
        University of Louisville\\
        Louisville, KY 40292, USA
}


\maketitle



\vspace{-0.4in}
\begin{abstract}
  Is it possible to make statistical inference broadly accessible to
  non-statisticians without sacrificing mathematical rigor or
  inference quality? This paper describes BayesDB, a probabilistic
  programming platform that aims to enable users to query the probable
  implications of their data as directly as SQL databases enable them
  to query the data itself.  This paper focuses on four aspects of
  BayesDB: (i) BQL, an SQL-like query language for Bayesian data
  analysis, that answers queries by averaging over an implicit space
  of probabilistic models; (ii) techniques for implementing BQL using
  a broad class of multivariate probabilistic models; (iii) a
  semi-parametric Bayesian model-builder that auomatically builds
  ensembles of factorial mixture models to serve as baselines; and
  (iv) MML, a ``meta-modeling'' language for imposing qualitative
  constraints on the model-builder and combining baseline models with
  custom algorithmic and statistical models that can be implemented in
  external software. BayesDB is illustrated using three applications:
  cleaning and exploring a public database of Earth satellites;
  assessing the evidence for temporal dependence between macroeconomic
  indicators; and analyzing a salary survey.
\end{abstract}

\begin{keywords}
  Probabilistic programming, Bayesian inference, probabilistic
  databases, multivariate statistics, nonparametric Bayes, automatic
  machine learning
\end{keywords}

\thanks{{\bf Acknowledgements:} VKM would like to thank Alexey Radul,
  Feras Saad, and Taylor Campbell for helpful discussions and
  contributions to a prototype implementation. This research was
  supported by DARPA (under the XDATA and PPAML programs), IARPA
  (under research contract 2015-15061000003), the Office of Naval
  Research (under research contract N000141310333), the Army Research
  Office (under agreement number W911NF-13-1-0212), the Bill \&
  Melinda Gates Foundation, and gifts from Analog Devices and Google.}

\section{Introduction}

Is it possible to make statistical inference broadly accessible to
non-statisticians without sacrificing mathematical rigor or inference
quality? This paper describes BayesDB, a system that enables users to
query the probable implications of their data as directly as SQL
databases enable them to query the data itself. By combining ordinary
SQL with three new primitives --- {\tt SIMULATE}, {\tt INFER}, and
{\tt ESTIMATE} --- users of BayesDB can detect predictive
relationships between variables, retrieve statistically similar data
items, identify anomalous data points and variables, infer missing
values, and synthesize hypothetical subpopulations.  The default
modeling assumptions that BayesDB makes are suitable for a broad class
of problems \citep{crosscatjmlr, wasserman11}, but statisticians can
customize these assumptions when necessary. BayesDB also enables
domain experts that lack statistical expertise to perform qualitative
model checking \citep{gelmanEtal95} and encode simple forms of
qualitative prior knowledge.

BayesDB consists of four components, integrated into a single
probabilistic programming system:

\begin{enumerate}
\item The Bayesian Query Language (BQL), an SQL-like query language
  for Bayesian data analysis. BQL programs can solve a broad class of
  data analysis problems using statistically rigorous formulations of
  cleaning, exploration, confirmatory analysis, and predictive
  modeling. BQL defines these primitive operations for these workflows
  in terms of Bayesian model averaging over results from an implicit
  set of multivariate probabilistic models.

\item A mathematical interface that enables a broad class of
  multivariate probabilistic models, called generative population
  models, to be used to implement BQL. According to this interface, a
  data generating process defined over a fixed set of variables is
  represented by (i) an infinite array of random realizations of the
  process, including any observed data, and (ii) algorithms for
  simulating from arbitrary conditional distributions and calculating
  arbitrary conditional densities. This interface permits many
  statistical operations to be implemented once, independent of the
  specific models that will be used to apply these operations in the
  context of a particular data table.

\item The BayesDB Meta-modeling Language (MML), a minimal
  probabilistic programming language. MML includes constructs that
  enable statisticians to integrate custom statistical models ---
  including arbitrary algorithmic models contained in external
  software --- with the output of a broad class of Bayesian model
  building techniques. MML also includes constructs for specifying
  qualitative dependence and independence constraints.

\item A hierarchical, semi-parametric Bayesian ``meta-model'' that
  automatically builds ensembles of generalized mixture models from
  database tables. These ensembles serve as baseline data generators
  that BQL can use for data cleaning, initial exploration, and other
  routine applications.

\end{enumerate}

This design insulates end users from most statistical considerations.
Queries are posed in a qualitative probabilistic programming language
for Bayesian data analysis that hides the details of probabilistic
modeling and inference. Baseline models can be built automatically and
customized by statisticians when necessary. All models can be
critically assessed and qualitatively validated via predictive checks
that compare synthetic rows (generated via BQL's {\tt SIMULATE}
operation) with rows from the original data. Instead of hypothesis
testing, dependencies between variables are obtained via Bayesian
model selection.

BayesDB is ``Bayesian'' in two ways:

\begin{enumerate}

\item In BQL, the objects of inference are rows, and the underlying
  probability model forms a ``prior'' probability distribution on the
  fields of these rows. This is then constrained by row-specific
  observations to create a posterior distribution of field
  values. Without this prior, it would be impossible to simulate rows
  or infer missing values from partial observations.

\item In MML, the default meta-model is Bayesian in that it assigns a
  prior probability to a very broad class of probabilistic models and
  narrows down on probable models via Bayesian inference. This prior
  is unusual in that it encodes a state of ignorance rather than a
  strong inductive constraint. MML also provides instructions for
  augmenting this prior to incorporate qualitative and quantitative
  domain knowledge.

\end{enumerate}

In practice, it is useful to use BQL for Bayesian queries against
models built using non-Bayesian or only partially Bayesian
techniques. For example, MML supports composing the default meta-model
with modeling techniques specified in external code that need not be
Bayesian. However, the default is to be Bayesian for both model
building and query interpretation, as this ensures the broadest
applicability of the results.

This paper focuses on the technical details of BQL, the data generator
interface, the meta-model, and the MML. It also illustrates the
capabilities of BayesDB using three applications: cleaning and
exploring a public database of Earth satellites, discovering
relationships in measurements of macroeconomic development of
countries, and analyzing salary survey data. Empirical results are
based on a prototype implementation that embeds BQL into {\tt
  sqlite3}, a lightweight, open-source, in-memory database.

\subsection{A conceptual illustration}

This section illustrates data analysis using the MML and BQL on a
synthetic example based on analysis of electronic health records. SQL
databases make it easy to load data from disk and run queries that
filter and retrieve the contents. The first step in using BayesDB is
to load data that describes a statistical (sub)population into a
table, with one row per member of the population, and one column per
variable:

\begin{quote}
{\tt CREATE POPULATION patients WITH DATA FROM patients.csv; \\
\\
SELECT age, has\_heart\_disease FROM patients WHERE age > 30 LIMIT 3;\\

\begin{tabular}{c|c}
age & has\_heart\_disease \\
\hline
66 & ??? \\
44 & yes \\
31 & ???
\end{tabular}
}
\end{quote}

Once data has been loaded, a {\em population schema} needs to be
specified. This schema specifies the statistical characteristics of
each example. For example, whether it is categorical or numerical, and if
it is categorical, how many outcomes are there and how is each outcome
represented. After an initial schema has been specified --- using a
mix of automatic inference and manual specification --- the schema can
be customized using instructions in the Meta-modeling Language (MML).

\begin{quote}
{\tt GUESS POPULATION SCHEMA FOR patients;\\
\\
ALTER POPULATION SCHEMA FOR patients\\
  SET DATATYPE FOR num\_hosp\_visits TO COUNT;\\

CREATE DEFAULT METAMODEL FOR patients;\\
ALTER METAMODEL FOR patients ENSURE will\_readmit DEPENDENT ON dialysis;\\
\\ALTER METAMODEL FOR patients \\
MODEL infarction GIVEN gender, age, weight, height, cholesterol, bp USING CUSTOM MODEL FROM infarction\_regression.py;
}
\end{quote}

One distinctive feature of MML is that it includes instructions for
{\em qualitative probabilistic programming}. These instructions
control the behavior of the automatic modeling machinery in the MML
runtime. In this example, these constraints include the assertion of a
dependence between the presence of a chronic kidney condition and
future hospital readmissions. They also include the specification of a
custom statistical model for the {\tt infarction} variable,
illustrating one way that discriminative and non-probabilistic
approaches to inference can be integrated into BayesDB.

The next step is to use the MML to build an ensemble of
general-purpose models for the data, subject to the specified constraints:

\begin{quote}
{\tt INITIALIZE 100 MODELS FOR patients; \\
ANALYZE patients FOR 3 HOURS CHECKPOINT EVERY 10 MINUTES;
}
\end{quote}

Each of these 100 models is a {\em generative population model} (GPM)
that represents the joint distribution on all possible measurements of
an infinite population with the given population schema. These models
are initially drawn accordingt to a broad prior probability
distribution over a large hypothesis space of possible GPMs. Until the
observed data has been analyzed, BQL will thus report broad
uncertainty for all its query responses.

Once the models are sufficiently adapted to the data, it is possible
to query its probable implications. The following query quantifies
over columns, rather than rows, and retrieves the probability of a
marginal dependence between three (arbitrary) variables and {\tt height}:

\begin{quote}
{\tt ESTIMATE COLUMN NAME, PROBABILITY OF DEPENDENCE WITH height\\
FROM COLUMNS OF patients LIMIT 3; \\

{\em
\begin{tabular}{c|c}
column name & p( dep.~with height )\\
\hline
height & 1.0 \\
infarction & 0.08 \\
gender & 0.99
\end{tabular}
}}
\end{quote}

Point predictions can be accessed by using the {\tt INFER}
instruction, a natural generalization of {\tt SELECT} from SQL:

\begin{quote}
{\tt INFER age, has\_heart\_disease FROM patients \\WHERE age > 30 WITH CONFIDENCE 0.8 LIMIT 3;\\

{\em
\begin{tabular}{c|c}
age & has\_heart\_disease \\
\hline
66 & {\em yes} \\
44 & yes \\
31 & ???
\end{tabular}
}}
\end{quote}

In this example, only one of the missing values could be inferred with
the specified confidence level. The probabilistic semantics of {\tt
  CONFIDENCE} will be discussed later in this paper.

BQL also makes it straightforward to generate synthetic sub-populations subject to a broad class of constraints:
\begin{quote}
{\tt
SIMULATE height, weight, blood\_pressure FROM patients 3 TIMES \\GIVEN gender = male AND age < 10\\

{\em \begin{tabular}{c|c|c}
height & weight & blood\_pressure \\
\hline
46 & 80 & 110 \\
38 & 60 & 80 \\
39 & 119 & 120
\end{tabular}
}}
\end{quote}

The {\tt SIMULATE} operator gives BQL users access to samples from the
posterior predictive distribution induced by the implicit underlying
set of models. This is directly useful for predictive modeling and
also decision-theoretic choice implemented using Monte Carlo
estimation of expected utility \citep{russelln03}. It also enables
predictive checking: samples from {\tt SIMULATE} can be compared to
the results returned by {\tt SELECT}. Finally, domain experts can use
{\tt SIMULATE} to scrutinize the implications of the underlying model
ensemble, both quantitatively and qualitatively.

\section{Example Analyses}

This section describes three applications of the current BayesDB
prototype:

\begin{enumerate}

\item Exploring and cleaning a public database of Earth satellites.

\item Assessing the evidence for dependencies between indicators of global poverty

\item Analyzing data from a salary survey.

\end{enumerate}

BQL and MML constructs are introduced via real-world uses; a
discussion of their formal interpretation is provided in later
sections.

\subsection{Exploring and cleaning a public database of Earth satellites}

The Union of Concerned Scientists maintains a database of ~1000 Earth
satellites. For the majority of satellites, it includes kinematic,
material, electrical, political, functional, and economic
characteristics, such as dry mass, launch date, orbit type, country of
operator, and purpose. Here we show a sequence of interactions with a
snapshot of this database using the {\tt bayeslite} implementation of
BayesDB.

\subsubsection{Inspecting the data.}

We start by loading the data and looking at a sample. This process
uses a combination of ordinary SQL and convenience functions built
into {\tt bayeslite}. The first step is to create a population from
the raw data:

\begin{quote}
{\tt CREATE POPULATION satellites FROM ucs$\_$database.csv}
\end{quote}

One natural query is to find the International Space
Station, a well-known satellite:

\begin{quote}
{\tt SELECT * FROM satellites WHERE Name LIKE 'International Space Station\%'}
\end{quote}

\begin{tabular}{|l|l|}
\hline
{\bf Variable} & {\bf Value} \\
\hline
\hline
Name&	International Space Station (ISS ...) \\
 Country\_of\_Operator&	Multinational \\
Operator\_Owner&	NASA/Multinational \\
Users&	Government \\
Purpose&	Scientific Research \\
Class\_of\_Orbit&	LEO \\
Type\_of\_Orbit&	Intermediate \\
Perigee\_km&	401\\
Apogee\_km&	422 \\
Eccentricity&	0.00155 \\
Period\_minutes&	92.8 \\
Launch\_Mass\_kg&	NaN \\
Dry\_Mass\_kg&	NaN \\
Power\_watts&	NaN \\
Date\_of\_Launch&	36119 \\
Anticipated\_Lifetime&	30 \\
Contractor&	Boeing Satellite Systems (prime)/Multinational \\
Country\_of\_Contractor&	Multinational \\
Launch\_Site&	Baikonur Cosmodrome \\
Launch\_Vehicle&	Proton \\
Source\_Used\_for\_Orbital\_Data&	www.satellitedebris.net 12/12 \\
longitude\_radians\_of\_geo&	NaN \\
Inclination\_radians&	0.9005899 \\
\hline
\end{tabular}

\begin{tabular}{c}\\\end{tabular}

This result row illustrates typical characteristics of real-world
databases such as heterogeneous data types and missing values.




\subsubsection{Building baseline models.}

Before exploring the implications of the data, it is necessary to
obtain a collection of probabilistic models. The next two MML
instructions produce a collection of 16 models, using roughly 4
minutes of analysis total.

\begin{quote}
{\tt INITIALIZE 16 MODELS FOR satellites; \\
ANALYZE satellites FOR 4 MINUTES WAIT;}
\end{quote}

Each of the 16 models is a separate GPM produced by an independent
Markov chain for approximate posterior sampling in the default
semi-parametric factorial mixture meta-model described earlier. This
number of models and amount of computation is typical for the
exploratory analyses done with our prototype implementation; this is
sufficient for roughly 100 full sweeps of all latent variables.


\subsubsection{Answering hypotheticals.}

The satellites database should in principle inform the answers to a
broad class of hypothetical or ``what if?'' questions. For example,
consider the following question:

\begin{quote}

  {\em Suppose you receive a report indicating the presence of a
    previously undetected satellite in geosynchronous orbit with a dry
    mass of 500 kilograms.  What countries are most likely to have
    launched it, and what are its likely purposes?}

\end{quote}

Answering this question requires knowledge of satellite engineering,
orbital mechanics, and the geopolitics of the satellite industry. It
is straightforward to answer this question using BQL. The key step is
to generate a synthetic population of satellites that reflect the
given constraints:

\begin{quote}
{\tt SIMULATE country\_of\_operator, purpose FROM satellites\\
GIVEN Class\_of\_orbit = GEO, Dry\_mass\_kg = 500
LIMIT 1000;
}
\end{quote}

\begin{figure}
\includegraphics[width=6in]{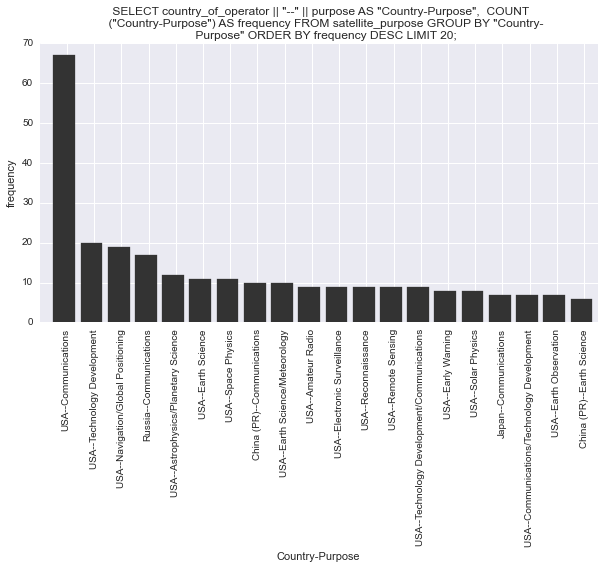}
\caption{{\bf The most probable countries and purposes of a satellite
    with a 500 kilogram dry mass in geosynchronous orbit.} See main
  text for discussion.}
\label{fig:satellites-simulate}
\end{figure}

Figure~\ref{fig:satellites-simulate} shows the results of a simple
aggregation of these results, counting the marginal frequencies of
various countries and purposes and sorting accordingly. The most
probable explanation, carrying roughly 25\% of the probability mass,
is that it is a communications satellite launched by the USA. It is
also plausible that it might have been launched other major space
powers such as Russia or China, and that it might have a military
purpose.

The satellites data are too sparse and ambiguous for frequency
counting to be a viable alternative. Consider an approach based on
finding satellites that match the discrete {\tt GEO} constraint and
are within some ad-hoc tolerance around the observed dry mass:

\begin{quote}
{\tt SELECT country\_of\_operator, purpose, Class\_of\_orbit, Dry\_mass\_kg \\
FROM satellites \\
WHERE Class\_of\_orbit = "GEO" AND Dry\_Mass\_kg BETWEEN 400 AND 600;
}
\end{quote}

This SQL query returns just 2 satellites, both Indian:

\begin{tabular}{c}\\\end{tabular}

\begin{tabular}{|c|c|c|c|c|}
\hline
 & {\bf Country\_of\_Operator} & {\bf Purpose} & {\bf Class\_of\_Orbit}&{\bf Dry\_Mass\_kg}\\
\hline
\hline
0&	India&	Communications&	GEO&	559\\
1&	India&	Meteorology&	GEO&	500\\
\hline
\end{tabular}

\begin{tabular}{c}\\\end{tabular}

Presuming our intuition about satellite mass is flawed, we might issue
another query to look at a broader range of satellites:
\begin{quote}
{\tt SELECT country\_of\_operator, purpose, Class\_of\_orbit, Dry\_mass\_kg \\
FROM satellites\\
WHERE Class\_of\_orbit = 'GEO'
AND Dry\_Mass\_kg BETWEEN 300 AND 700}
\end{quote}

\begin{tabular}{c}\\\end{tabular}

The results still do not give any real insight into the likely purpose
of this satellite:

\begin{tabular}{c}\\\end{tabular}

\begin{tabular}{|c|c|c|c|c|}
\hline
&\bf Country\_of\_Operator&\bf Purpose&\bf Class\_of\_Orbit&\bf Dry\_Mass\_kg\\
\hline
\hline
0&Malaysia&Communications&GEO&650\\
1&Israel&Communications&GEO&646\\
2&Luxembourg&Communications&GEO&700\\
3&Russia&Communications&GEO&620\\
4&China (PR)&Earth Science&GEO&620\\
5&China (PR)&Earth Science&GEO&620\\
6&China (PR)&Earth Science&GEO&620\\
7&India&Communications&GEO&559\\
8&India&Navigation&GEO&614\\
9&India&Meteorology&GEO&500\\
10&Malaysia&Communications&GEO&650\\
11&Multinational&Earth Science/Meteorology&GEO&320\\
12&United Kingdom&Communications&GEO&660\\
13&Norway&Communications&GEO&646\\
\hline
\end{tabular}

\begin{tabular}{c}\\\end{tabular}

Without deep expertise in satellites, and significant expertise in
statistics, it is difficult to know whether or not these results can
be trusted. How does the set of satellites vary as the thresholds on
{\tt Dry\_Mass\_kg} are adjusted? How locally representative and
comprehensive is the coverage afforded by the data? Are there
indirect, multivariate dependencies that ought to be taken into
account, to determine which satellites are most similar?  How should
existing satellites be weighted to make an appropriate weighted sample
against which to calculate frequencies? In fact, small modifications
to the tolerance on {\tt Dry\_Mass\_kg} yield large changes in the
result set.

\subsubsection{Identifying predictive relationships between variables.}

A key exploratory task is to identify those variables in the database
that appear to predict one another. This is closely related to the key
confirmatory analysis question of assessing the evidence for a
predictive relationship between any two particular variables.

To quantify the evidence for (or against) a predictive relationship
between two pairs of variables, BQL relies on information theory. The
notion of dependence between two variables A and B is taken to be
mutual information; the amount of evidence for dependence is then the
probability that the mutual information between A and B is nonzero.
If the population models are obtained by posterior inference in a
meta-model --- as is the case with MML --- then this probability
approximates the posterior probability (or strength of evidence) that
the mutual information is nonzero.

\begin{quote}
{\tt ESTIMATE DEPENDENCE PROBABILITY FROM PAIRWISE COLUMNS OF satellites;}
\end{quote}

\begin{figure}
\begin{center}
\begin{subfigure}[b]{4in}
\includegraphics[width=4in]{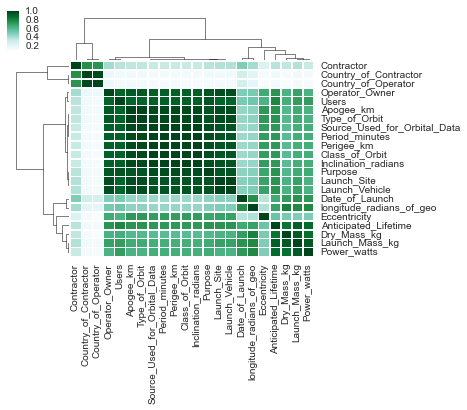}\\
\caption{A heatmap depicting the pairwise probabilities of
  dependence between all pairs of variables. 
  The rows and columns are both permuted according to a
  single ordering obtained via agglomerative hierarchical clustering
  to highlight multivariate interactions.}
\end{subfigure}
\begin{subfigure}[b]{4in}
\includegraphics[width=4in]{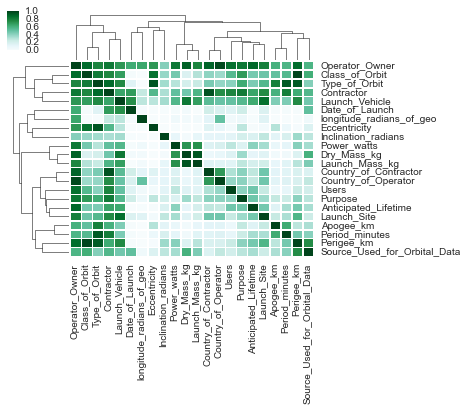}
\caption{The pairwise
  correlation matrix; note that many causal relationships are not
  detected by simple correlations. See main text for discussion.}
\end{subfigure}
\end{center}
\caption{{\bf Detecting predictive relationships in the satellites
    database.}}
\label{fig:satellites-depprob}
\end{figure}

Figure~\ref{fig:satellites-depprob} shows the results from this
query. There are several groups of variables with high probability of
mutual interdependence. For example, we see a definite block of
geopolitically related variables, such as the country of contractor \&
operator, the contractor's identity, and the location of the satellite
(if it is in geosynchronous orbit). The kinematic variables describing
orbits, such as perigee, apogee, period, and orbit class, are also
shown as strongly interdependent. A domain expert with sufficiently
confident domain knowledge can use this overview of the predictive
relationships to assess the value of the data and the validity of the
baseline models.

It is also instructive to compare the heatmap of pairwise dependence
probabilities with alternatives from
statistics. Figure~\ref{fig:satellites-depprob} also shows heatmap
that results from datatype-appropriate measures of correlation. The
results from correlation are sufficiently noisy that it would be
difficult to trust inferences from techniques that use correlation to
select variables. Furthermore, the most causally unambiguous
relationships, such as the kinematic constraints relating perigee,
apogee, and orbital period, not detected by correlation.

\subsubsection{Detecting multivariate anomalies.}

Another key aspect of exploratory analysis is identifying anomalous
values, including both (univariate) outliers and multivariate
anomalies. Anomalies can arise due to errors in data acquisition, bugs
in upstream preprocessing software (including binning of continuous
variables or translating between different discrete outcomes), and
runtime failures. Anomalies can also arise due to genuine surprises or
changes in the external environment.

Using BQL, multivariate anomalies can be detected by assessing the
predictive probability density of each measurement, and ordering from
least to most probable. Here we illustrate this using a simple
example: ordering the geosynchronous satellites according to the
probability of their recorded orbital period:

\begin{quote}
{\tt ESTIMATE name, class\_of\_orbit, period\_minutes AS TAU, \\
PREDICTIVE PROBABILITY OF period\_minutes AS "Pr[TAU]" \\
FROM satellites\\ ORDER BY ``Pr[TAU]'' ASCENDING LIMIT 10
}
\end{quote}

This BQL query produces the following table of results:

\begin{tabular}{c}\\\end{tabular}
\hspace{-0.75in}\begin{tabular}{|c|c|c|c|c|}
\hline
& {\bf Name} & {\bf Class\_of\_Orbit} & {\bf TAU} & {\bf Pr[TAU]} \\
\hline
\hline
0&	AEHF-3 (Advanced Extremely High Frequency sate...	&GEO&	1306.29&	0.001279\\
1&	AEHF-2 (Advanced Extremely High Frequency sate...	&GEO&	1306.29&	0.001292\\
2&	DSP 20 (USA 149) (Defense Support Program)	&GEO&	142.08&	0.002656\\
3&	Intelsat 903&	GEO&	1436.16&	0.003239\\
4&	BSAT-3B	&GEO	&1365.61&	0.003440\\
5&	Intelsat 902&	GEO	&1436.10	&0.003492\\
6&	SDS III-6 (Satellite Data System) NRO L-27, Gr...	&GEO	&14.36	&0.003811\\
7&	Advanced Orion 6 (NRO L-15, USA 237)	&GEO	&23.94	&0.003938\\
8&	SDS III-7 (Satellite Data System) NRO L-38, Dr...&	GEO&	23.94	&0.003938\\
9&	QZS-1 (Quazi-Zenith Satellite System, Michibiki)	&GEO&	1436.00	&0.004446\\
\hline
\end{tabular}

\begin{tabular}{c}\\\end{tabular}

Recall that a geosynchronous orbit should take 24 hours or 1440
minutes. Rows 7 and 8 appear to be unit conversion errors (hours
rather than minutes). Rows 2 and 6 appear to be decimal placement
errors. Note that row 2 is not an outlier: some satellites have an
orbital period of roughly two hours. It is only anomalous in the
context of the other variables that are probably predictive of orbital
period, such as orbit class.

\subsubsection{Inferring missing values.}

A key application of predictive modeling is inferring point
predictions for missing measurements. This can be necessary for
cleaning data before downstream processing. It can also be of
intrinsic interest, e.g. in classification problems. The satellites
database has many missing values. Here we show an {\tt INFER} query
that infers missing orbit types and returns both a point estimate and
the confidence in that point estimate:

\begin{quote}
{\tt INFER EXPLICIT \\
anticipated\_lifetime, perigee\_km, period\_minutes, class\_of\_orbit,\\
PREDICT type\_of\_orbit AS inferred\_orbit\_type CONFIDENCE inferred\_orbit\_type\_conf \\
FROM satellites \\
WHERE type\_of\_orbit IS NULL;}
\end{quote}

\begin{figure}
\includegraphics[width=6in]{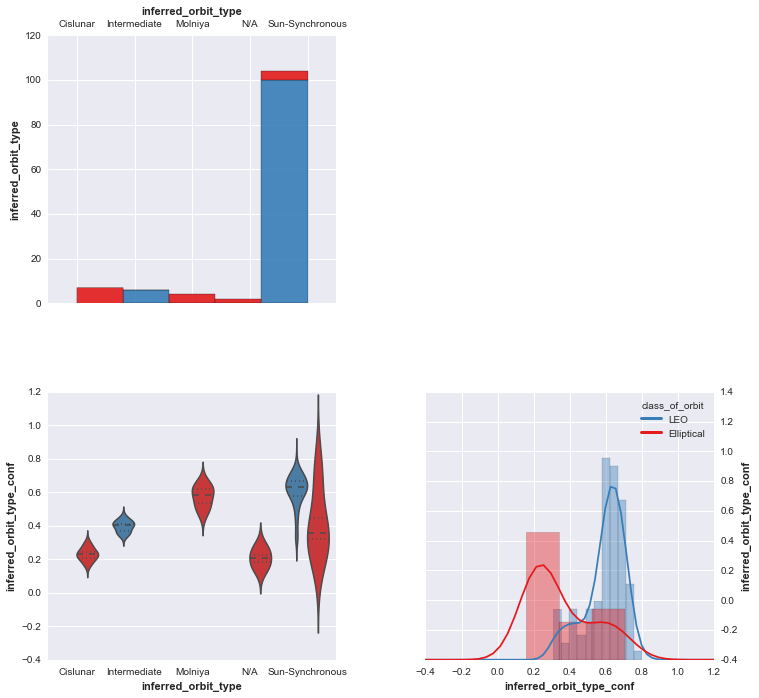}
\caption{{\bf A visualization of inferred point estimates for {\tt
      type\_of\_orbit} and the confidence in those point estimates.}
  See main text for discussion.}
\label{fig:satellites-infer}
\end{figure}

This form of {\tt INFER} uses the {\tt EXPLICIT} modifier that exposes
both predicted values and their associated confidence levels to be
included in the output. Figure~\ref{fig:satellites-infer} shows a
visualization of the results. The panel on the bottom left shows that
the confidence depends on the orbit class and on the predicted value
for the inferred orbit type. For example, there is typically moderate
to high confidence for the orbit type of {\tt LEO} satellites --- and
high confidence (but some variability in confidence) for those with
{\tt Sun-Synchronous} orbits. Satellites with {\tt Elliptical} orbits
may be assigned a {\tt Sun-Synchronous} type with moderate confidence,
but for other target labels confidence is generally lower. After
examining the overall distribution on confidences, it can be natural
to filter {\tt INFER} results based on a manually specified confidence
threshold. Note that many standard techniques for imputation from
statistics correspond to {\tt INFER ... WITH CONFIDENCE 0}.

\subsubsection{Integrating a kinematic model for elliptical orbits.}

Can we improve over the baseline models by integrating causal
knowledge about satellites? MML can be used to compose GPMs built by
the default model builder with algorithmic and/or statistical models
specified as external software. Here we integrate a simple model for
elliptical orbits:

\begin{quote}
{\tt
ALTER METAMODEL FOR satellites \\
MODEL perigee\_km, apogee\_km GIVEN period\_minutes, eccentricity \\
USING CUSTOM MODEL FROM stochastic\_kepler.py;\\
ANALYZE FOREIGN PREDICTORS FOR 1 MINUTE;
}
\end{quote}

The underlying foreign predictor implements Kepler's laws:

\begin{flalign*}
R_{min} & = \tau^{\frac{2}{3}} (1.0 - \epsilon) - R_{GEO} \\
R_{max} & = \tau^{\frac{2}{3}} (1.0 + \epsilon) - R_{GEO} \\
X_{(r^*, {\tt apogee\_km})} & \sim N(R_{min}, \sigma^2) \\
X_{(r^*, {\tt perigee\_km})} & \sim N(R_{max}, \sigma^2)
\end{flalign*}

Here, $\epsilon$ is set via the {\tt eccentricity} measurement for row
$r^*$, and $\tau$ is set via the {\tt period\_minutes}
measurement. $R_{GEO}$ is a fixed constant inside the foreign
predictor representing the radius of the Earth in kilometers. $\sigma$
is a parameter that determines the noise and is set when the variable
subset for the foreign predictor instantiation is {\tt ANALYZE}d. Note
that foreign predictors are essentially GPMs and accordingly must
implement generic simulate(...) and logpdf(...)
methods. For algorithmic forward models with numerical outputs, MML
provides a default wrapper that uses importance sampling with
resampling to approximately generate conditional samples and estimate
marginal densities. This is how Kepler's laws --- in the form of a
forward simulation --- are turned into a generative model for
kinematic variables that can be conditionally simulated in arbitrary
directions.

\begin{figure}
\begin{center}
{\tt SIMULATE period\_minutes, apogee\_km FROM satellites\_kepler LIMIT 100;}

\includegraphics[width=7in]{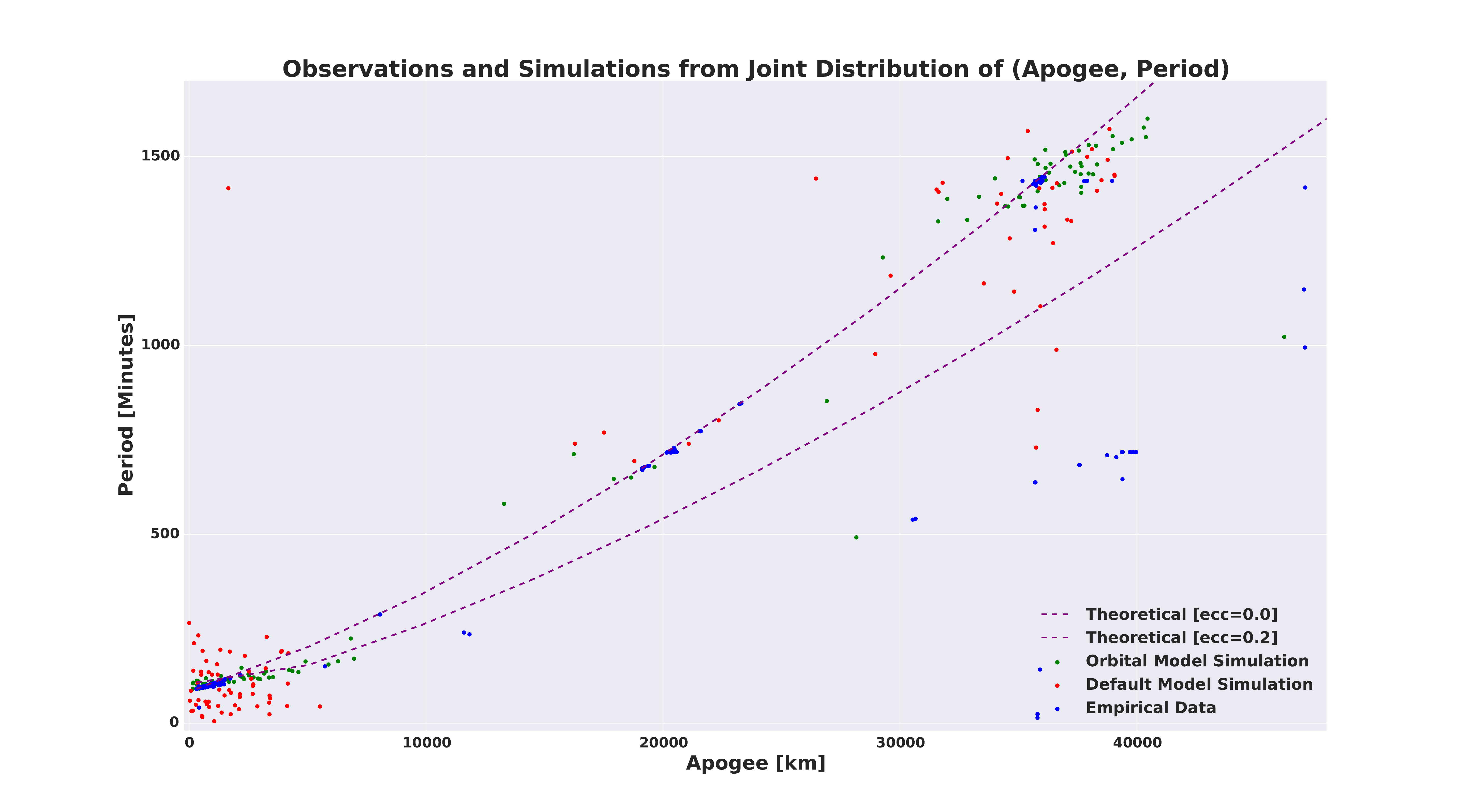}
\end{center}
\caption{{\bf Integrating empirical baseline models with a noisy
    version of Kepler's laws.} Neither the empirical approach nor the
  analytical approach is universally dominant in terms of
  accuracy. See main text for discussion.}
\label{fig:satellites-kepler-joint}
\end{figure}

\begin{figure}
\begin{center}
{\tt SIMULATE perigee\_km, apogee\_km FROM satellites\_kepler ASSUMING period\_minutes = 1436 LIMIT 100;}

\includegraphics[width=6in]{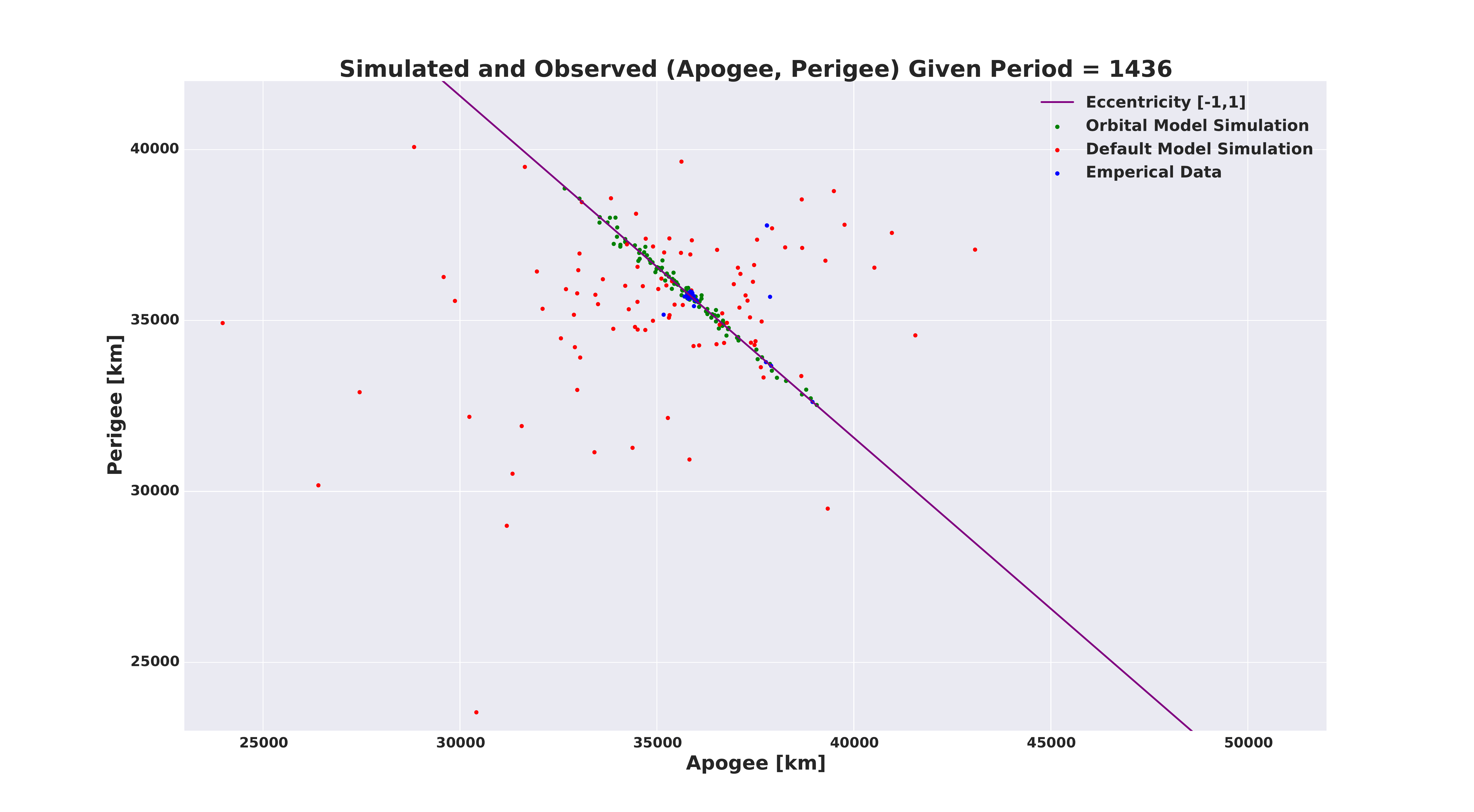}

\includegraphics[width=6in]{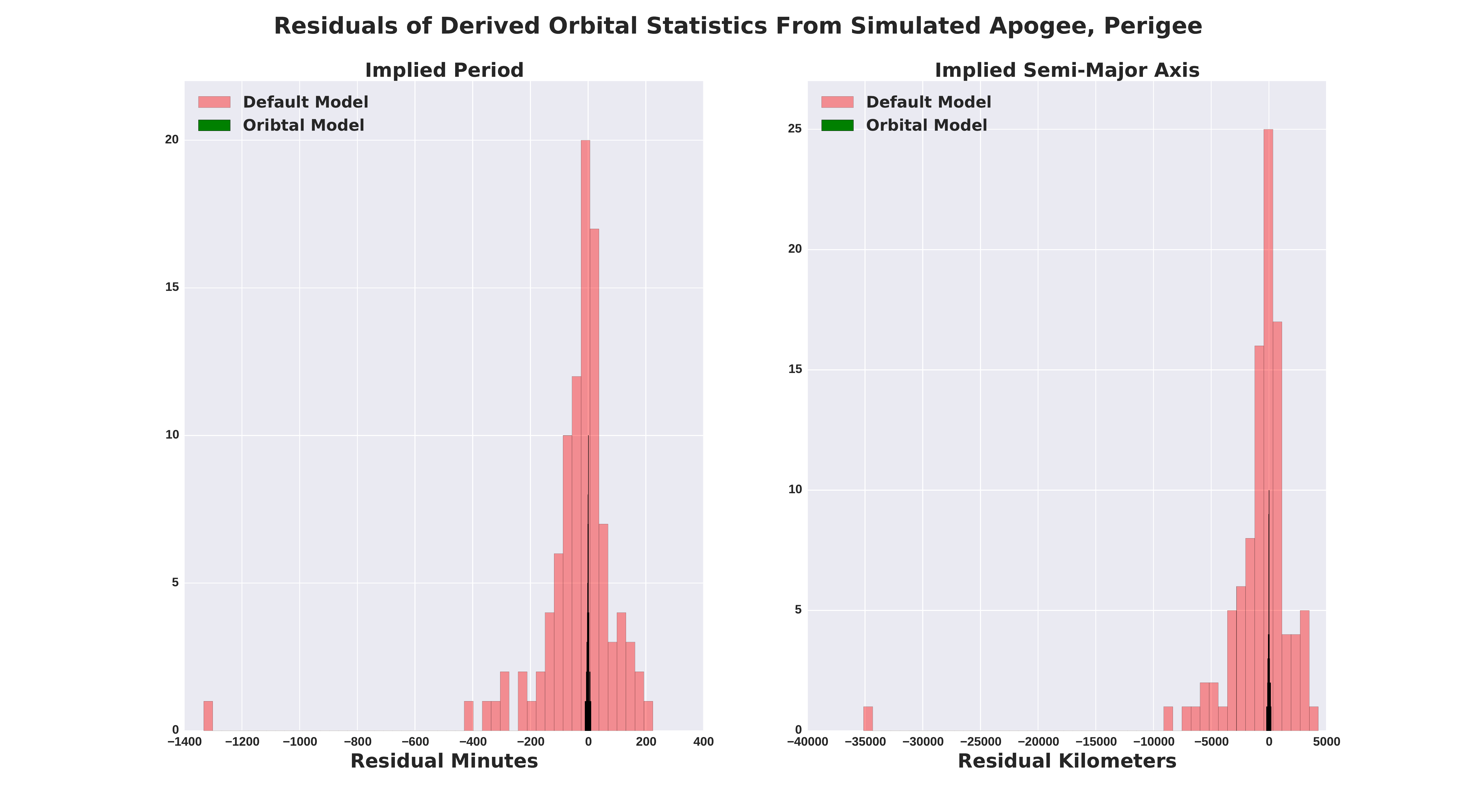}
\end{center}
\caption{{\bf Comparing conditional predictions of an empirical model
    with Kepler's laws.} See main text for discussion.}
\label{fig:satellites-kepler-conditional}
\end{figure}

Figure~\ref{fig:satellites-kepler-joint} and
Figure~\ref{fig:satellites-kepler-conditional} show the results. A
detailed discussion of the relative merits of empirical versus
analytical modeling is beyond the scope of this paper. However, it is
clear that neither the empirical approach nor the analytical approach
is universally dominant. The empirical approach is able to correctly
locate the empirical probability mass --- including multiple modes ---
but underfits. The orbital mechanics approach yields inferences that
are typically in much tighter accord with the kinematic data. This is
unsurprising: these are the patterns of covariation that led to the
development of quantitative models and ``hard'' natural
sciences. However, there are many satellites for which Kepler's laws
are not in accord with the data. This reflects many factors, including
data quality errors as well as legitimate gaps between idealized
mathematical laws and fine-grained empirical records of real-world
phenomena. For example, at the time Kepler's laws were formulated,
orbiting bodies lacked engines.

\subsubsection{Combing random forests, causal models, and nonparametric Bayes.}

Because MML supports model composition, it is straightforward to build
hybrid models that integrate techniques from subfields of machine
learning that might seem to be in conflict. Figure \ref{fig:MMLKepler} shows the transcript of
a complete MML session that builds such a hybrid model. Random forests
are used to classify orbits into types; Kepler's Laws are used to
relate period, perigee, and apogee; and the default semi-parametric
Bayesian meta-model is used for all remaining variables (with two
variables coming with overridden datatypes). 

\begin{figure}
\centering
\begin{minipage}{4in}
\small
\belowcaptionskip=-10pt
\begin{verbatim}
CREATE POPULATION satellites
  FROM ucs_satellites.csv

CREATE METAMODEL sat_keplers ON satellites
  USING composer(
  random_forest (
    Type_of_Orbit (CATEGORICAL)
      GIVEN Apogee_km, Perigee_km,
            Eccentricity, Period_minutes,
            Launch_Mass_kg, Power_watts,
            Anticipated_Lifetime,
 Class_of_orbit
  ),
  foreign_model (
    source = 'keplers_laws.py',
    Period_Minutes (NUMERICAL)
      GIVEN Perigee_km, Apogee_km
  ),
  default_metamodel (
    Country_of_Operator CATEGORICAL,
    Inclination_radians NUMERICAL
  )
);

INITIALIZE 16 MODELS FOR satellites;
ANALYZE satellites FOR 4 MINUTES;
\end{verbatim}
\end{minipage}
\caption{A complete MML session that builds a hybrid model integrating techniques from subfields of machine learning that might seem to be in conflict.}
\label{fig:MMLKepler}
\end{figure}

Later sections of this paper explain how these three modeling
approaches are combined to answer individual BQL queries.

\subsection{Assessing the evidence for dependencies between indicators of global poverty}

In the early 21st century it is widely believed that resolving extreme
poverty around the world will be accomplished by empowering
individuals to resolve their poverty. Governments and NGOs encourage
this process through a variety of interventions, many of them
combining material assistance with policy changes. In principle,
policies should be driven by quantitative data-driven understanding of
international economic development. In practice, international
economic data is sparse, unreliable, and highly aggregated. These data
limitations create substantial obstacles to understanding the
situational context of successful or unsuccessful interventions and
policies.

The ``Gapminder'' data set, collected and curated by Hans Rosling at the
Karolinska Institutet, is the most well known and extensive sets of
longitudinal global developmental indicators. Representing over 500
indicators, 400 countries, and 500 years of data, it covers the
colonial era, industrial revolution, socio-political upheavals around
the world in the 20th century, and the first decade of the
21st. Containing over 2 million observations, the data has been used
as the basis for a compelling set of data animations and the most
widely viewed TED talk on statistics.


To date, analysis of this data has been minimal, as it requires
intensive preprocessing and cleaning. Different analytical
methdologies require different approaches to imputation and variable
selection; as a result, results from different teams are difficult to
compare. Here we show how to explore the data with BayesDB and assess
the evidence for predictive relationships between different
macroeconomic measures of development.

\subsubsection{Exploring the Data with SQL}

\begin{figure}
\centering
\begin{tabular}{|c|c|c|c|c|}
\hline
{\bf observation\_count} & {\bf indicator\_count} & {\bf
                                                    country\_count} & {\bf
                                                                     year\_count}
                  & {\bf
                                                                     coverage}\\

\hline
\hline
2082193 & 464 & 405 & 364 & 0.03044 \\
\hline
\end{tabular}

\begin{subfigure}[b]{\linewidth}
\includegraphics[width=\linewidth]{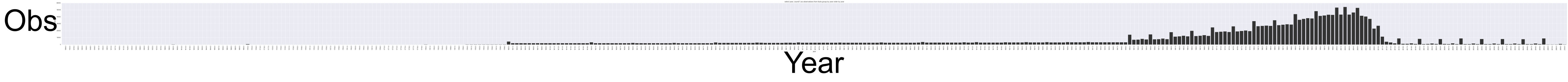}
\caption{\textbf{Observation volume by year, 1086-2100.}\ Note that the
most complete data is for only recent history.}
\label{fig:gapminder-by-year}
\end{subfigure}

\begin{subfigure}[b]{\linewidth}
\includegraphics[width=\linewidth]{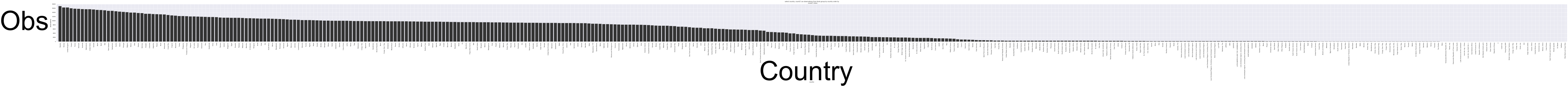}
\caption{\textbf{Observation volume by country.}\ Note that the data
  becomes very limited for some countries, and includes countries that
  no longer exist.}
\label{fig:gapminder-by-country}
\end{subfigure}

\begin{subfigure}[b]{\linewidth}
\includegraphics[width=\linewidth]{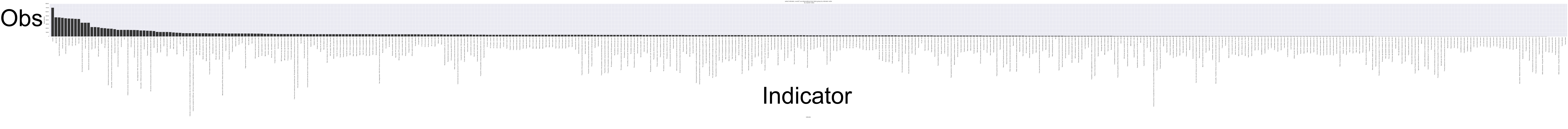}
\caption{\textbf{Observation volume by indicator.}\ Note that some
  indicators are much more complete or extensive in terms of year and
  country than others.}
\label{fig:gapminder-by-indicator}
\end{subfigure}

\caption{{\bf Gapminder data volume measures.} The dataset contains
  longitudinal records of $\sim$500 macroeconomic variables for
  $\sim$300 countries, spanning a $\sim$100 year period. However,
  roughly 97\% of the data is missing.}
\label{fig:gapminder-data}
\end{figure}

The raw form of the data is $\sim$500 Excel spreadsheets, each
containing longitudinal data for $\sim$300 countries over $\sim$100
years. However, the dataset only contains $\sim$2 million
observations, i.e. 97\% of the data is
missing. Figure~\ref{fig:gapminder-data} shows key indicators of the
data around size, missing records, and the relationship between data
availability and countries, records, and years. The primary data is
mmodeled in SQL as a “fact” table structure. This
relatively-normalized representation easily models the sparse matrix
and allows us to use a combination of SQL and Python data science
tools to craft our population structures.

The histograms in Figure~\ref{fig:gapminder-data} show the breadth and
also the variability of the data. The histogram by year in
Figure~\ref{fig:gapminder-by-year} shows that data is complete for
only recent history, and in fact that some predicted data continues
into the future, and that data for some indicators is only available
every 10 years. The histogram by country in
Figure~\ref{fig:gapminder-by-year} shows that data availability varies
by country (the most described country is Sweden), that many countries
have reasonably complete data, but that there is a long tail of
countries with sparse data, including countries that no longer exist
and with inconsistent or disputed
naming. Figure~\ref{fig:gapminder-by-indicator} shows that there is
also a variance by indicators, because different measurements are
collected by different agencies with different expectations and data
policies.

The data has already been subject to extensive visualization and
descriptive analytics by the Gapminder project. This paper focuses on
the use of MML to model the data and BQL to query its probable
implications.

\subsubsection{Detecting Basic and Longitudinal Dependence}


Our analysis focuses on the 53 variables with most complete data for
the years 1999-2008. It is straightforward to create an ensemble of
models for this subset:

\begin{quote}
{\tt
GUESS POPULATION SCHEMA FOR dense\_gapminder; \\
INITIALIZE 64 MODELS FOR dense\_gapminder; \\
ANALYZE todo FOR 300 MINUTES WAIT;
}
\end{quote}

\begin{figure}
\centering
\begin{subfigure}[b]{.9\linewidth}
\includegraphics[width=5in]{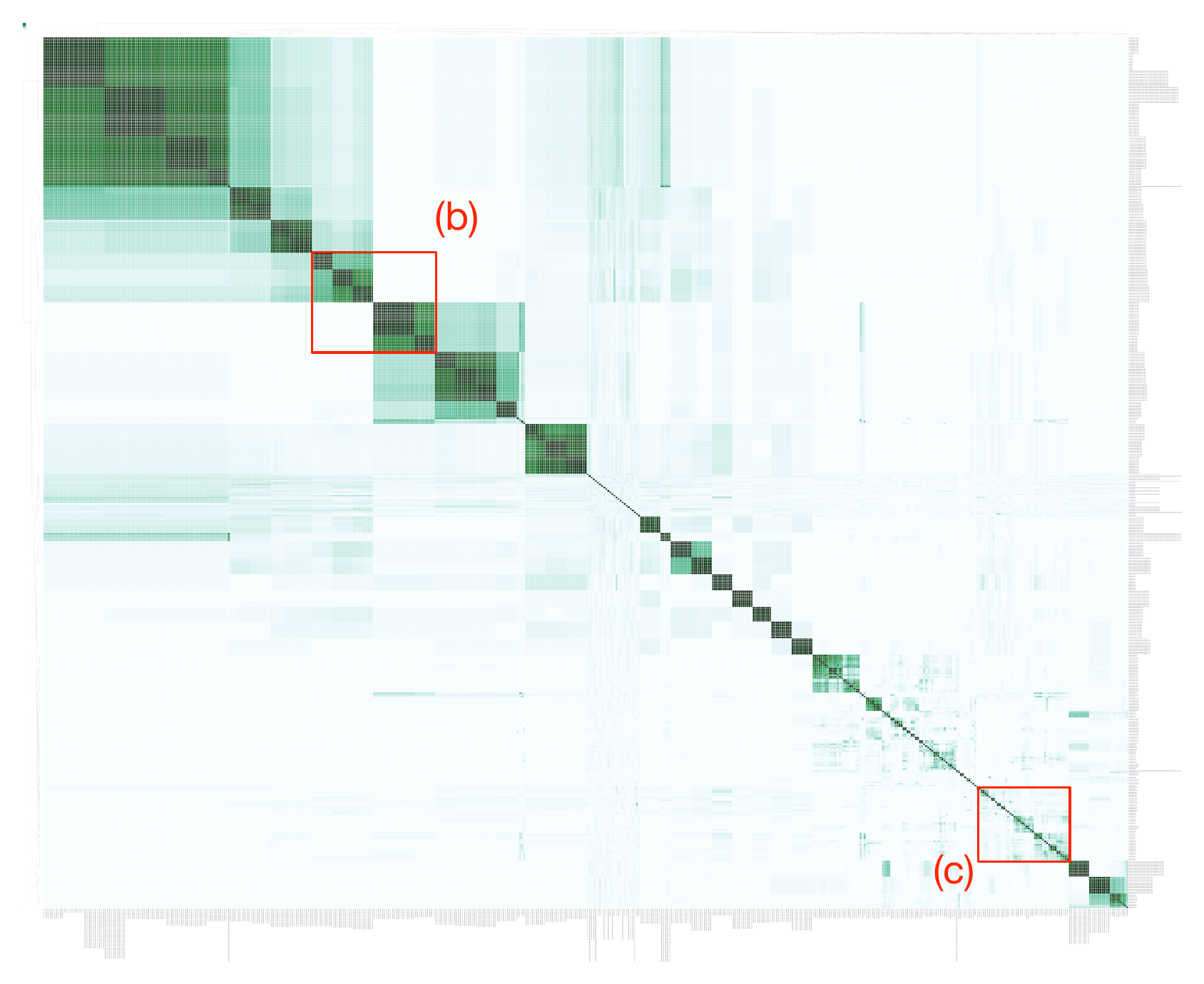}
\caption{Probability of dependence heatmap for 40 indicators over 10 years.}
\end{subfigure}

\begin{subfigure}[b]{.45\linewidth}
\includegraphics[width=2.75in]{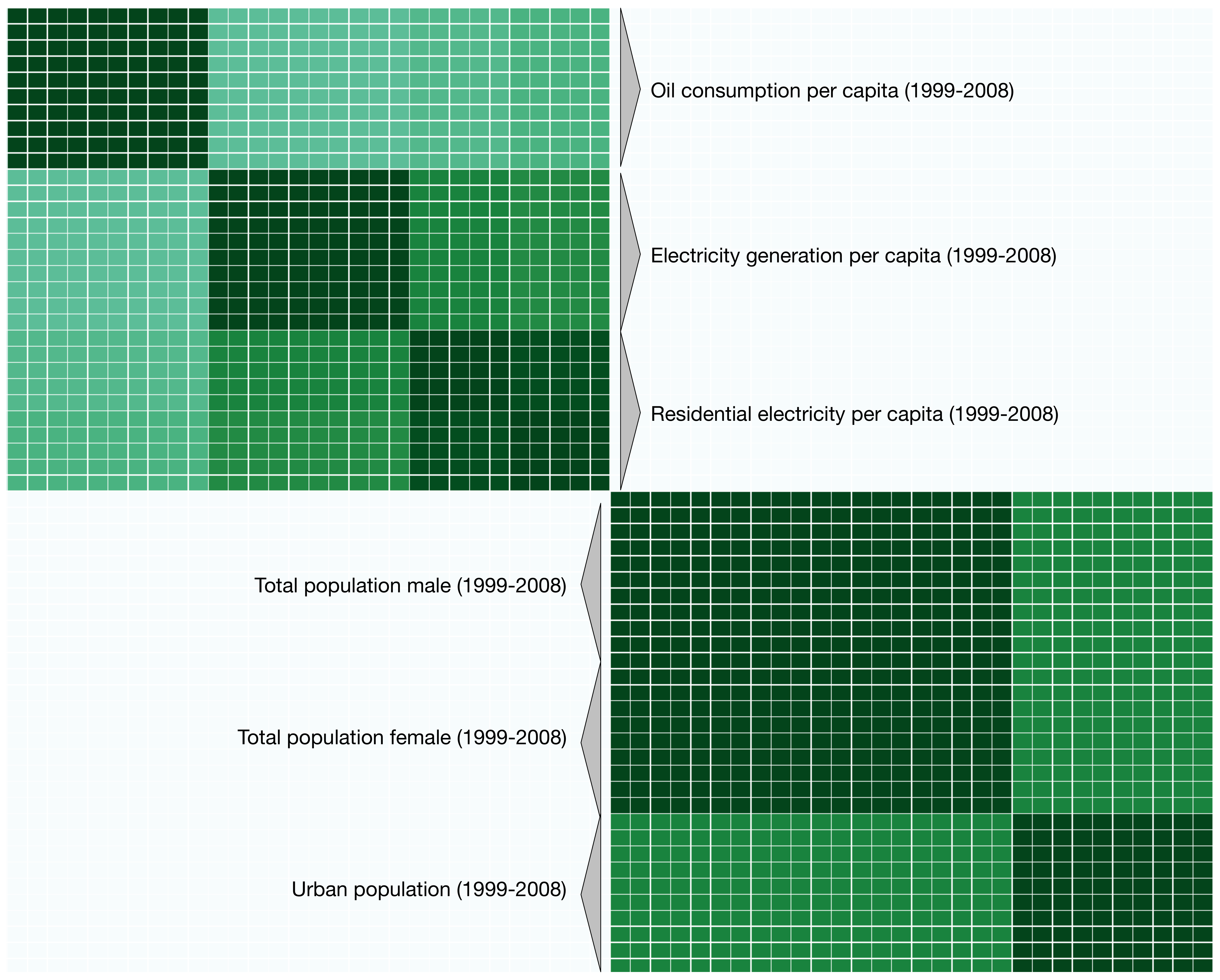}
\caption{Indicators form 10 year runs, with a sharp break from totals
  to per capita.}
\label{fig:gapminder-probdep-zoom1}
\end{subfigure}
\hfill
\begin{subfigure}[b]{.45\linewidth}
\includegraphics[width=2.75in]{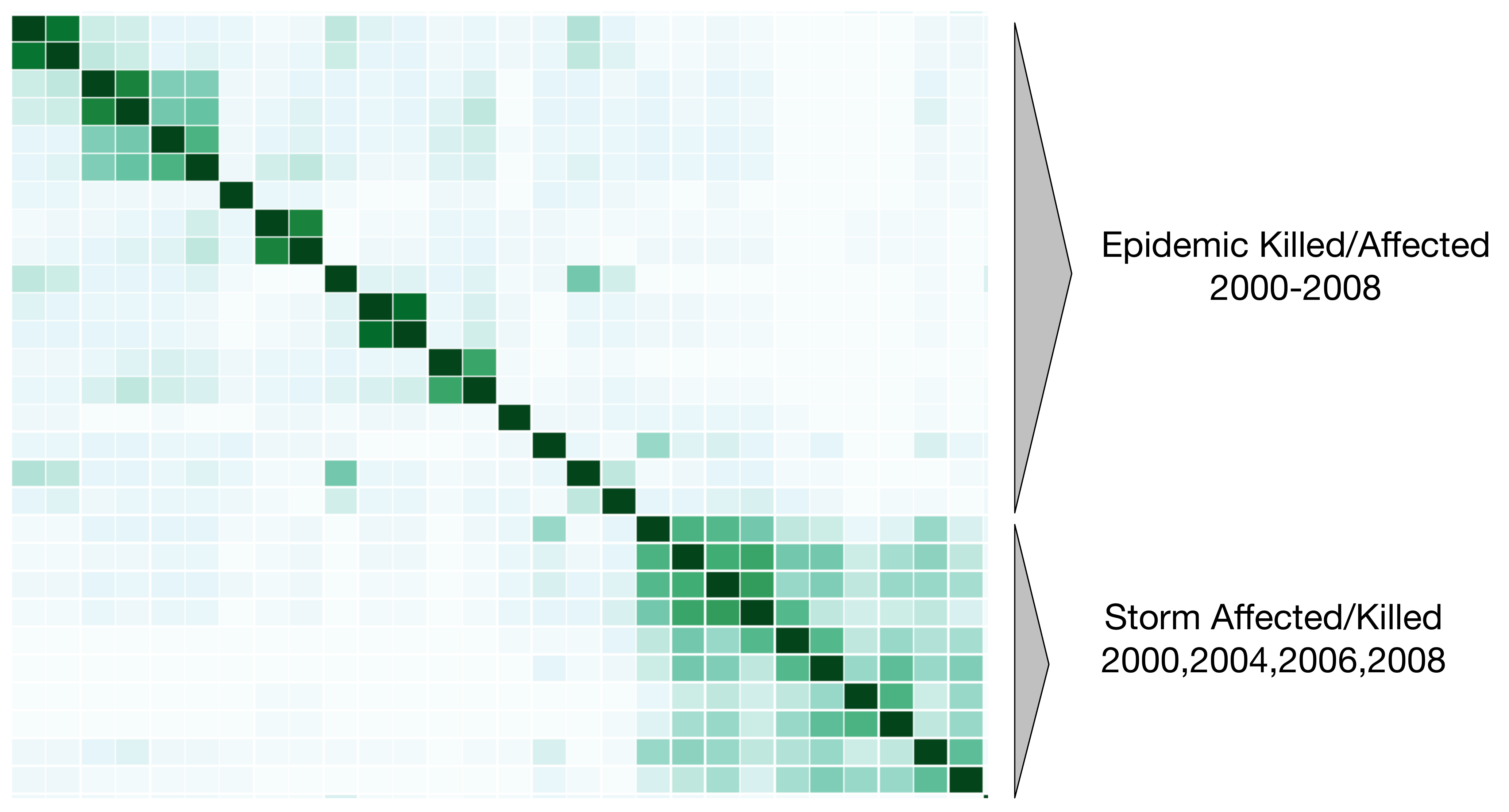}
\caption{Natural disaster indicators cluster but do not have strong
  year-to-year dependence.}
\label{fig:gapminder-probdep-zoom2}
\end{subfigure}

\caption{\textbf{Probability of dependence heatmap for the Gapminder
    data.}\ BayesDB detects temporal dependence within some indicators
  but not others, as well as dependence between some indicators (but
  not others). See main text for discussion.}
\label{fig:gapminder-probdep}
\end{figure}

The probability of dependence heatmap that results is shown in
Figure~\ref{fig:gapminder-probdep}. Indicators such as total
population and urban percentage form blocks containing their values
for all 10 years contained in the dataset. This shows that the default
GPM was able to extract the temporal dependence in these
indicators. In other cases, such as measurements of the number of
people killed in floods, the year to year dependence is much
weaker. The heatmap also shows dependence between indicators, such as
the block in the top right corner combining indicators of stress,
urbanization, and fertility rate. Finally, it segregates data
according to type sof indicators, as can be seen in
Figure~\ref{fig:gapminder-probdep-zoom1} where there is a sharp break
from total measurements to per-capital.

\begin{figure}
  \includegraphics[width=\linewidth]{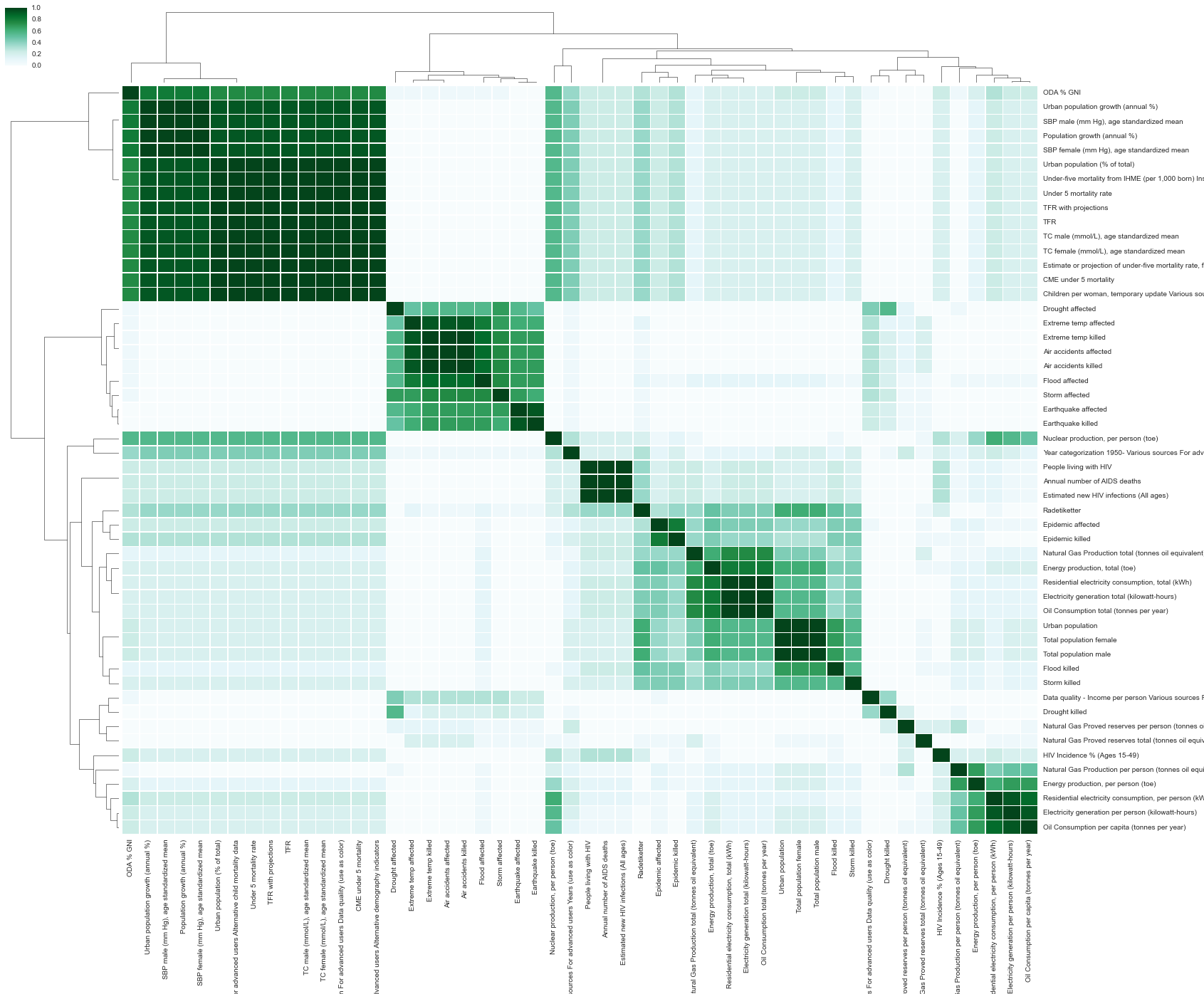}
  \caption{\textbf{Probability of dependence for 40 indicators in
      2002.}\ Of particular note are the blocks for population growth
    rates, natural disasters, HIV, total energy usage, and per capita
    energy usage.}
  \label{fig:gapminder2002}
\end{figure}

If we analyze just the data for the year 2002, using 32 models for 3
minutes, we get a heatmap that highlights the dependence and
independence between indicators. Figure~\ref{fig:gapminder2002} shows
the details.

\subsubsection{Measuring the Similarity of Countries}

\begin{figure}
%
\begin{subfigure}[t]{3in}
\includegraphics[width=3in]{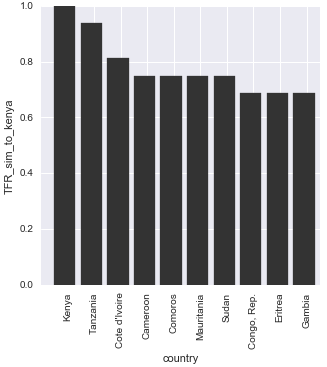}
\caption{Countries enumerated in decending order of similarity to
  Kenya on Total Fertility Rate.}
\end{subfigure}
\hspace{.25in}
\begin{subfigure}[t]{3in}
\includegraphics[width=3in]{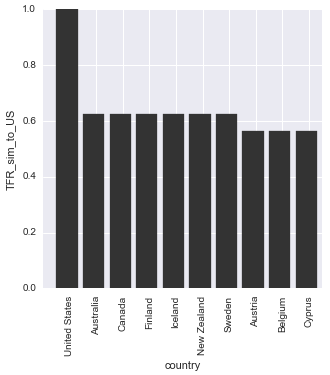}
\caption{Countries enumerated in descending order of similarity to the
United States on Total Fertility Rate.}
\end{subfigure}

\begin{subfigure}[t]{3in}
\includegraphics[width=3in]{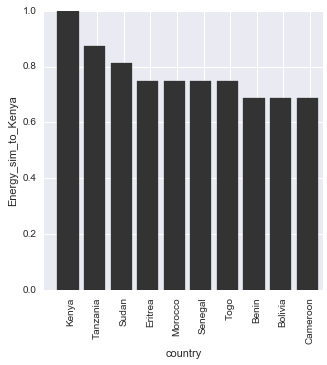}
\caption{Countries enumerated in decending order of similarity to
  Kenya on per capita energy production.}
\end{subfigure}
\hspace{.25in}
\begin{subfigure}[t]{3in}
\includegraphics[width=3in]{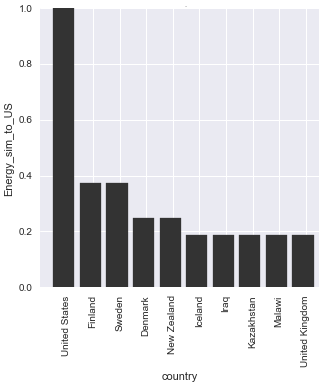}
\caption{Countries enumerated in descending order of similarity to the
United States on per capita energy production.}
\end{subfigure}
\caption{\textbf{Examining the similarity of countries.} Kenya and the
  United States are similar to different countries, and the similarity
  structure with respect to per capita energy production and total
  fertility rate are significantly different.}
\label{fig:gapminder-similarity}
\end{figure}

In order to help with the delivery of international aid and the design
and analysis of interventions, decision makers often want a richer
understanding of the similarities between countries. With BQL we can
formulate these queries in general or against specific
attributes. Figure~\ref{fig:gapminder-similarity} shows country
similarities for different indicators. As expected, changing the
indicator of interest can produce a very different similarity
structure. Analyses that presume a single global similarity measure
cannot pick up this context-specific structure.

The authors are involved in an ongoing research partnership with the
Bill and Melinda Gates Foundation aimed at integrating the Gapminder
data with other relevant sources, including qualitative knowledge from
domain eperts, and using it to drive empirically grounded policy and
aid interventions.

\subsection{Analyzing a salary survey}

Surveys are a common source of multivariate data and a potentially
appealing application for BayesDB. Here we show a preliminary analysis
of a web-administered anonymous salary survey. Participants shared
their compensation details along with information about their title,
years of service, acheivements, employer, and geography.

\subsection{Controlling Models with Qualitative Assumptions}

This salary population provides an instructive example of applying
qualitiative assumptions to a model. In this case, the first analysis
of compensation data finds that geographic location (state, region) is
not a factor in compensation. Domain experts suggest that is
implausible, that cost of living and the competitive market in
different cities is a significant factor in compensation of the survey
participants. The following code can be used to apply this qualitative
assumption:

\begin{quote}
{\tt ALTER METAMODEL FOR salary ENSURE total, equity, base, bonus
  DEPENDENT ON state;}
\end {quote}

\begin{figure}
\begin{subfigure}[b]{3in}
\includegraphics[width=3in]{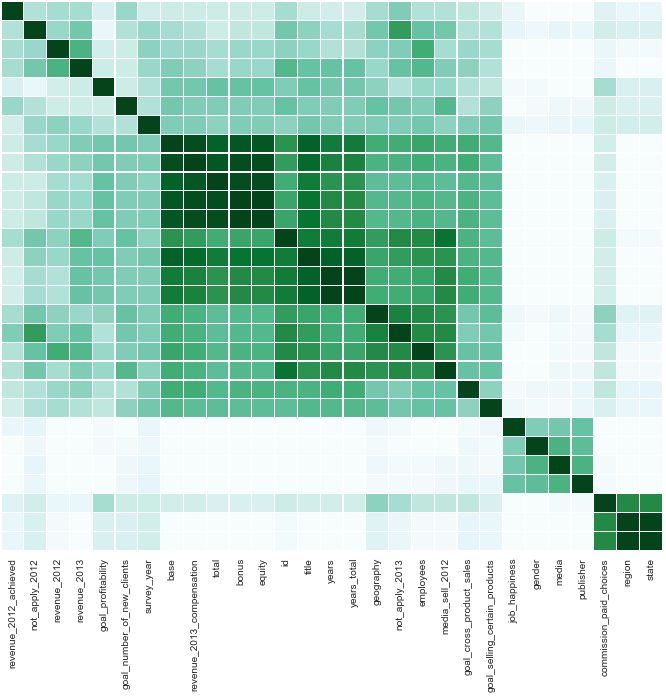}
\label{fig:salary-depprob}
\caption{Probability of dependence between columns with default
  metamodel settings.}
\end{subfigure}
\hspace{.25in}
\begin{subfigure}[b]{3in}
\includegraphics[width=3in]{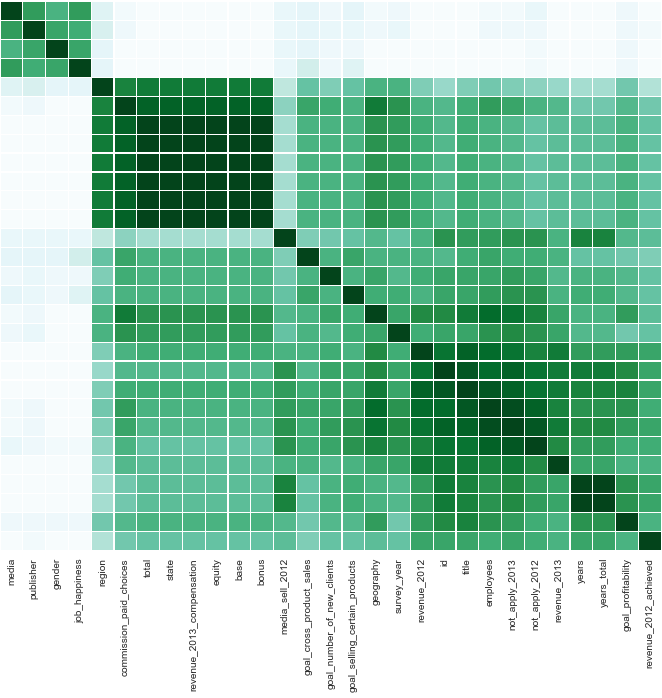}
\caption{Probability of dependence between columns when
  \texttt{state}\ is dependent on salary columns.}
\end{subfigure}

\caption{\textbf{Probability of dependence heat maps with and without
    dependence assertions in MML.}\ Note that in the second figure
  stronger dependencies were resolved overall.}
\label{fig:salary-state-depprob}
\end{figure}

Without asserting the dependence, state is inferred to be dependent on
region and independent of performance.  After asserting a qualitative
constraint, the probability of dependence heat map changes. Not only
are the squares implied by that depencence colored to 1.0, but other
columns have re-aligned in their modeling. In particular, given this
assumption, there appears to be more evidence of dependence between
the 2012 and 2013 measures and core indicators such as years in the
job, bonus, the presence of an equity stake, etc. Also, there appears
to be less evidence that job title impacts the key compensation
variables.

\section{The Bayesian Query Language}

The Bayesian Query Language (BQL) formalizes Bayesian data analysis
without exposing the end user to model parameters, priors, and
posteriors. This section describes the statistical operations that are
implemented by the core BQL instruction set. It also describes the
modeling formalism that is used to implement BQL.

\subsection{Generative Population Models}

BQL programs are executed against a weighted collection of {\em
  generative population models} (GPMs). At present, GPMs can be built
in two ways:

\begin{enumerate}

\item Specified directly as external software libraries.

\item Inferred from data via probabilistic inference in a meta-model
  written in BayesDB's Meta-modeling Language.

\end{enumerate}

GPMs can respond to queries about the joint distribution of the
underlying data generating process as a whole or about the predictive
distribution for a specific member of the population. The population
can be thought of as a table, where individual members are specified
by row indexes.

Each GPM induces a random table with a finite number of columns and an
infinite number of rows, where each cell contains a random
variable. BQL treats each BayesDB generator as a model of the data
generating process underlying its associated table of observations.
It is sometimes useful to query a GPM about hypothetical members of
the population. This can be performed by using a row whose index $r^*$
may not be associated with any actual member; this can be guaranteed
by generating a unique row index.

Each GPM is described by a {\em schema} $\mathcal{S}$ that must be
compatible with the population schema for the population to which it
is being applied. This schema is a tuple containing
$(\texttt{typed-outputs}, \texttt{typed-inputs}, \texttt{body})$. The
\texttt{typed-outputs}\ component specifies the column indexes and
statistical types of each column that the data generator will be
responsible for producing. The \texttt{typed-inputs}\ component specifies
the indexes and statistical types of each column that the data
generator can read from. The \texttt{body}\ is an opaque binary that
contains any GPM-specific configuration information, such as a
probabilistic program.

Mathematically, the internals of a GPM $\mathcal{G} = (\Theta,
\mathcal{Z}, O)$ consists of three parts:

\begin{enumerate}

\item Measurement-specific latent variables $\mathcal{Z} =
  \cup z_{(r,c)}$.

  There may be overlap between the latent variables for different
  measurements. If a GPM cannot track dependencies internally --- or
  if it is based on a model class in which all measurements are
  coupled --- then $z_{(r,c)} = \mathcal{Z}$.

\item Population-level latent variables $\Theta$.

  These are all latent variables that remain well-defined in the
  absence of all measurements. Examples include hyper-parameters and
  mixture component parameters.

\item Observations $O = \{(r_i, c_i, x_{(r_i, c_i)})\}$.

  These correspond to the observed measurements.

\end{enumerate}

For example, a naive Bayesian GPM lacks any measurement-specific
variables, i.e.  $z_{(r,c)} = \emptyset$, and is completely
characterized by a single vector of parameters $\Theta =
\vec{\theta_c}$ for the probability models for each column. A finite
mixture GPM would have $z_{(r,c)} = \{z_r\}$ be the cluster assignment
for each row, and have $\Theta = \{ \theta_{(c,l)} | l \in
\mathcal{Z}\}$ be the component model parameters for each cluster.

Generative population models are required to satisfy the following
conditional independence constraint:

$$
x_{(r,c)} | \Theta, z_{(r,c)} \bigCI x_{(r',c')} | \Theta, z_{(r',c')}\ \mathrm{unless}\ (r,c)=(r',c')
$$

Note in particular that the observations $O$ need not be conditioned
on directly, given $\Theta$ and $z_{(r,c)}$. This formalizes the
requirement that the dependencies between the measurements in the
population are completely mediated by the population-level latent
variables and all relevant measurement-specific latent variables. In
general, no other independence constraints are enforced by the
interface. GPMs can thus be built around dense, highly-coupled model
families such as low-dimensional latent spaces and convolutional
neural networks.

\subsubsection{An interface to generative population models}

A GPM must implement the following interface:

\begin{enumerate}
\item $\mathcal{G}$ = $\{\Theta, \mathcal{Z}\}$ = initialize( schema = $\mathcal{S}$ )

  Initialize a data generator with the given schema and return the
  resulting data generator $\mathcal{G}$. It ensures that storage has
  been allocated for the random variables $\Theta$ and $\mathcal{Z}$,
  storing the global latent variables and the local latent variables,
  respectively.

\item $\vec{s_i}$ = simulate($\mathcal{G}$, givens = $\{(r_j,
    c_j, x_{(r,c_j)})\}$, targets = $\{ r_k, c_k \}$, $N$)

  Generate $N$ sampled values $\{\vec{s_i}\}$ from the specified
  distribution:

$$
\{ \vec{s_i} \} \sim \{ X_{(r_k, c_k)} \} | \{ X_{(r_j, c_j)} =
x_{(r_j, c_j)} \}, \Theta, \{z_{r,c} | (r,c) \in \{r_j, c_j\} \cup
\{r_k, c_k\} \}
$$

The set of valid distributions includes all finite-dimensional joint
distributions obtainable by conditioning on the values of arbitrary
measurements and marginalizing over another arbitrary set.

\item $\log p$ = logpdf($\mathcal{G}$, givens = $\{(r_j, c_j,
    x_{(r_j,c_j)})\}$, query = $\{(r_k, c_k, q_{(r,c_k)}) \}$)

  Evaluate the log probability density of the specified
  conditional/marginal distribution at a target point:
$$\log p = \log p( \{X_{(r_k, c_k)} = q_{(r_k,c_k)} \} | \{ X_{(r_j, c_j)} = x_{(r_j, c_j)} \}, \Theta, \{z_{r,c} | (r,c) \in \{r_j, c_j\} \cup
\{r_k, c_k\} \})$$

\item $d$ = kl-divergence-given-G($\mathcal{G}$, measurements$\_$A = $\{(r_i, c^a_i)\}$, measurements$\_$B = $\{(r_j, c^b_j)\}$, conditions$\_$C = $\{(r_k, c^c_k, x^k)\}$)

  This estimates the KL divergence of the set of measurements $A$ from
  the set of measurements $B$, conditioned on the given constraints
  $C$. KL calculations are central to model-independent data
  analysis. For example, to detect predictive relationships, it
  suffices to check for non-zero mutual information, which can be
  reduced to calculating the KL between the joint distribution over
  two variables and the product of the marginals.

  It is included in the GPM interface because that allows a GPM
  implementer to supply an optimized implementation. Where such an
  implementation is not available, the KL can be estimated via simple
  Monte Carlo estimation:

\hspace{-0.5in}\begin{flalign*}
\mathrm{D}^\mathcal{G}_{KL}(\{X_{(r_i, c^a_i)}\}, \{X_{(r_j, c^b_j)}\})
= & \sum_{\{x_i\} \in dom(\{X_{(r_i, c^a_i)}\})} p\left( \{X_{(r_i, c^a_i)}\} = \{x_i\} \left| \mathcal{G}\right.\right)
log\left( \frac{ p\left( \{X_{(r_j, c^b_j)}\} = \{x_i\} \left| \mathcal{G}\right.\right)}
{ p\left( \{X_{(r_i, c^a_i)}\} = \{x_i\} \left| \mathcal{G}\right.\right)}
\right) \\
\approx & \sum_{\{x_i\}^k} log\left( \frac{p\left( \{X_{(r_j, c^b_j)}\} = \{x_i\}^k \left| \mathcal{G}\right.\right) }
{ p\left( \{X_{(r_i, c^a_i)}\} = \{x_i\}^k) \left| \mathcal{G}\right.\right)}\right)\\
& \mathrm{with} \{x_i\}^k \sim \{X_{(r_i, c^a_i)}\}\\
\end{flalign*}

\end{enumerate}

This interface is intentionally quite general. It needs to support an
open set of primitives for Bayesian data analysis. This paper focuses
on the subset of this interface where all measurements come from the
same row. All the BQL operations used in this paper can be reduced to
explicit invocations of $simulate$, $logpdf$, and to Monte Carlo
estimates of Kullback-Leibler divergences implemented in terms
them. Some GPMs can significantly optimize some of these operations
relative to Monte Carlo baselines; such optimizations are likely to be
important in practice but are beyond the scope of this paper.

\subsubsection{Weighted collections of generative population models.}

BQL is executed against a weighted collection of GPMs $\mathcal{M}$:
$$
\mathcal{M} = \{(w_i, \mathcal{G}_i)\}
$$
In principle, these collections can include GPMs drawn from different
model classes. The weights are treated as prior probabilities. This
paper focuses on the case where the GPMs come from a single
meta-model, each produced by independent runs of a single Markov chain
for posterior inference in the meta-model given all available
measurements. In this case, assigning unit weights to all models $w_i
= 1$ results in BQL queries based on a Monte Carlo approximation to
Bayesian model averaging.

\subsection{Core instructions: {\tt SIMULATE}, {\tt ESTIMATE}, and
  {\tt INFER}}

Data analysis workflows in BQL are built around three core classes of
statistical operations:

\begin{enumerate}

\item Generating samples from predictive probability distributions,
  including both completions of existing rows in a data table as well
  as predictive distributions over hypothetical rows.

\item Estimating predictive probability densities and approximating
  derived information-theoretic quantities.

\item Summarizing multi-modal probability distributions with single
  values.

\end{enumerate}

These capabilities are exposed via three basic extensions to SQL that
each combine results from individual GPMs in different ways. They can
be composed with ordinary SQL to solve a broad range of data analysis
tasks:

\begin{enumerate}

\item {\em Detecting predictive relationships between variables}: {\tt
    ESTIMATE COLUMN PROBABILITY OF DEPENDENCE WITH {\em ...} }

  This yields an estimate of the marginal probability of dependence
  between the specified columns. This is equivalent to the probability
  that the mutual information between those two variables is nonzero,
  integrating over the weighted collection of GPMs that BayesDB
  maintains. If the GPMs are produced by an asymptotically consistent
  estimator of the joint distribution, then these probabilities will
  reflect non-linear, heteroscedastic, or context-specific
  dependencies that statistical aggregates (such as correlation or
  linear regression coefficients) will not.

\item {\em Regression, classification, semi-supervised learning, and
  imputation}: {\tt INFER {\em ...}}

  Each of these predictive modeling tasks requires filling in point
  estimates in different conditions. All of these can be viewed as
  special cases of {\tt INFER}, which handles arbitrary patterns of
  missing values and both continuous and discrete prediction targets.

\item {\em Anomaly/outlier detection}: {\tt ORDER BY PROBABILITY OF {\em col} ASCENDING LIMIT {\em k}}

  Anomalous cells can be found by predictive checking: identify the
  cells that are least likely under the inferred constellation of
  models. These may not be outliers in the standard univariate sense:
  the low probability may be due to interactions between several
  variables, even though each variable on its own is marginally
  typical.

\item {\em Retrieving similar rows}: {\tt ORDER BY SIMILARITY TO {\em row}}

  A broad class of structured search operations can be performed via
  information-theoretic measures of similarity between rows. These are
  useful in both data exploration and in more targeted search.

\item {\em Predictive model checking}: {\tt SIMULATE {\em ...}}

  By comparing aggregates from the output of {\tt SIMULATE} to the
  output of the analogous {\tt SELECT} statements, it is possible to
  do predictive checking without having to mention models, parameters,
  priors, or posteriors.

\end{enumerate}

\subsubsection{{\tt SIMULATE}: generating samples from arbitrary predictive distributions.}

The first, called {\tt SIMULATE}, provides a flexible interface to
sampling from posterior predictive distributions:

\begin{quote}
{\tt SIMULATE {\em target columns} FROM {\em population} [WHERE
    {\em row filter}] [ASSUMING {\em constraint}] [{\em k}
  TIMES]}
\end{quote}

The {\tt WHERE} clause is interpreted as a constraint to test against
all members of the population that have been observed so far. If it is
not supplied, the {\tt SIMULATE} command is executed against an
arbitrary as-yet-unobserved member of the population, i.e. a unique
row id from the standpoint of the GPM interface. The {\tt ASSUME}
clause is interpreted as an additional set of constraints to condition
each row on before generating the simulations.

For example, to generate a proxy dataset of two variables {\tt varA}
and {\tt varB}, one can write {\tt SIMULATE varA, varB FROM population
  100 TIMES}. As another example, consider the BQL command {\tt
  SIMULATE varA, varD FROM population 20 TIMES WHERE varB = True AND
  varC IS MISSING ASSUMING varC = 3.4}. This generates 20 simulated
values from $ p( {\tt varA}, {\tt varD} | {\tt varC} = 3.4 )$ for each
member of the population where {\tt varB} is equal to {\tt True} and
{\tt varC} is missing. This behavior may seem non-intuitive. For
example, a {\tt SIMULATE} invocation with {\tt WHERE true} returns
$Rk$ rows, where $R$ is the number of rows in the database and $k$ is
the number of output samples specified with the query. On the other
hand, {\tt WHERE false} yields an empty result set, always. However,
this semantics allows SQL aggregates to reduce the predictions for
individual source rows by grouping on the row identifiers.

To formally describe the meaning of simulate, we first introduce some
notation.  Let $w(\{x_{(r,c)} | c \in \mathcal{G}\})$ be the predicate
denoted by the {\tt WHERE} clause, i.e. $w(\cdot) = 1$ if the
predicate is satisfied and 0 otherwise. Let $R$ be the set of rows for
which there is at least one measurement, i.e. $R = \{r_i |
(r_i,\cdot,\cdot) \in O\}$, and let $W = \{r_i | w(\{{x_{(r_i,c)}} | c \in
\mathcal{G}\}) = 1\}$ be the set of rows that satisfy the {\tt WHERE}
clause's filter. If a {\tt WHERE} clause is not provided, then
$w(\{x_{(r,c)} | c \in \mathcal{G}\}) = 0$ for all existing rows $r
\in R$, and $W = \{ r^* \}$ be a set containing a single distinguished
row about which no measurements are known. Let $T = \{c_i\}$ be the
set of target columns, and let $A = (c_j, x_{(r,c_j)})$ be the set of
assumed equality constraints. Also let $TA = T \cup \{c | (c,\cdot)
\in A\}$ be the set of all columns referenced in the {\tt SIMULATE}
command.

For each $r^* \in W$, the {\tt SIMULATE} primitive produces a set of
$k$ returned realizations $S_{r^*} = \{ s_i \}$ of the following
generative process:
\begin{flalign*}
\mathcal{G}^i &\sim Discrete( \{ \mathcal{G}_j) ; w_j \} ) &&\hfill \\
s_i &\sim \{X_{r^*,c} | c \in T\} | \{X_{r^*,c'} = x_{c'} | (c',x_{c'}) \in A\},
\Theta^i, \{z^i_{(r^*,c_m)} | c_m \in TA\} &&\hfill
\end{flalign*}

This corresponds to choosing a GPM at random according to the
probabilities given by their weights and then generating $s_i$ from
the conditioned distribution in that model. If the models are equally
weighted, i.e. $w_i = 1$, and if all the GPMs are drawn from their
posterior distribution given the observations $p( \mathcal{G} | O )$,
then this procedure implements sampling from the Bayesian posterior
predictive distribution over the targets given all the observed data
plus the additional constraints from the {\tt ASSUME} clause:
\begin{flalign*}
p( \{X_{r^*,c} | c \in T\} | \{X_{r^*,c'} = x_{c'} | (c',x_{c'}) \in
A\}, O ) \\
\propto p( \{X_{r^*,c} | c \in T\} | \{X_{r^*,c'} = x_{c'} |
(c',x_{c'}) \in A\} | \mathcal{G} ) p( \mathcal{G} | O )
\end{flalign*}


\subsubsection{{\tt ESTIMATE}: approximating posterior averages.}

The second core BQL primitive, called {\tt ESTIMATE}, allows clients
to query the posterior expectations of stochastic functions that are
defined over the rows and the columns:

\begin{quote}
{\tt ESTIMATE {\em target properties} FROM [COLUMNS OF] {\em table} [WHERE {\em row/col filter}]}
\end{quote}

\noindent {\bf Row-wise estimands provided by BQL.} Consider the case
where the rows are being queried, i.e. {\tt COLUMNS OF} does not occur
in the query. Let $P = \{ f_i(x_{(r,c_i)}, \mathcal{G}) \}$ be the set
of properties whose values are requested. These properties can depend
on observed measurements as well as latent components of the GPM. Let
$w(\cdot)$ implement the {\tt WHERE} clause's filter, as with {\tt
  SIMULATE}. If a {\tt WHERE} clause is not provided, then
$w(\{x_{(r,c)} | c \in \mathcal{G}\}) = 0$ for all existing rows $r
\in R$.

Given these definitions, each row in the output of this class of {\tt
  ESTIMATE} invocations is defined as follows:
$$
\{e_i\}\ \mathrm{with}\ e_i = \sum_k w_k f_i( x_{(r,c_i)}, \mathcal{G}_k )
$$
The total set of returned rows is defined by the where clause:
$$
\{ \{e_i\}_r \}\ \mathrm{for}\ r \in W = \{r_i | w(\{{x_{(r_i,c)}} | c
\in \mathcal{G}\}) = 1\}
$$

\begin{enumerate}

\item $\log p$ = predictive-probability$(\mathcal{G}, {\tt row} = r, {\tt col} = c)$

  This estimand is denoted {\tt PREDICTIVE PROBABILITY OF {\em col}},
  and applied against an implicitly specified row, thus picking out a
  single measurement in the population. It can be implemented by
  delegation to the underlying GPM:

\begin{quote}
predictive-probability($\mathcal{G}, r, c)$) = logpdf($\mathcal{G}, \emptyset,  \{(r,c,x_{(r_j,c_j)}\ \mathrm{from}\ O_\mathcal{G})\}$)
\end{quote}

This can be used to identify outliers --- measurements that are
unlikely under their marginal distribution --- as well as anomalous
measurements that are marginally likely but unlikely given the other
measurements for the same row.

\item sim(a,b) = generative-similarity($\mathcal{G}$, context = $\{c_i\}$, rowA = $r^a$, rowB = $r^b$)

  Data analysts frequently want to retrieve rows from a table that are
  ``statistically similar'' to some pre-existing or hypothetical
  row. This is a key problem in data exploration. It is also useful
  when trying to explain surprising inference results or when trying
  to diagnose and repair data or inference quality issues. Many
  machine learning techniques treat similarity as a central primitive,
  and use a metric formulation of similarity as the basis for
  inductive generalization.

  Information theory provides appealing alternatives: measure
  similarity in terms of the amount of information one row contains
  about the values in another. This can be assessed against all
  variables or just against a ``context'' that is defined by a
  particular subset of variables. One approach, leading to a
  directional measure, is to measure the divergence of the
  distribution over values in one row from the distribution over
  values in another:

$$
Pr[ \mathrm{D}^\mathcal{G}_{KL}( \vec{x}_{\mathcal{G},r^a}|_{\{c_i\}} || \vec{x}_{\mathcal{G},r^b}|_{\{c_i\}} ) = 0]
$$

\end{enumerate}

\noindent {\bf Column-wise estimands provided by BQL.} Another use of
{\tt ESTIMATE} is to query properties of the columns, via {\tt
  ESTIMATE ... FROM COLUMNS OF ...}. Let $r^*$ be a distinguished
row about which no measurements are known, i.e. $(r^*, \cdot, \cdot)
\notin O$. Let $g_i(x_{(r^*,c)}, \mathcal{G})$ be a function of a set
of measurements from a fresh row and the underlying GPM. It is then
straightforward to define the set $G$ of values needed to check the
{\tt WHERE} filter, the set of columns $C_s$ that satisfy the filter,
and the set $E$ of returned values containing all the target
expressions for each satisfying column.

\begin{flalign*}
G & = \{g_c ( x_{(r^*,c)},  \{x_{(r,c)} | (r, c, \cdot) \in O\}, \mathcal{G}_k ) |c \in \mathcal{G}_k \}\\
C_s & = \{g | g \in G\ \mathrm{and}\ w(g) = 1\}\\
E & = \cup_t\ \{g_t ( x_{(r^*,c)}, \{(x_{(r,c)} | (r, c, \cdot) \in O\}, \mathcal{g}_k) |c \in C_s\}
\end{flalign*}


\begin{enumerate}
\item $p$ = marginal-dependence-prob$(\mathcal{G}, {\tt colA} = c_i, {\tt colB} = c_j)$

  This estimand characterizes the amount of evidence for the existence
  of a predictive relationship between the pair of variables $c_i$ and
  $c_j$. It is defined according to the information-theoretic
  definition of conditional independence:

$$
Pr[ X_{(r^*, c_i)} \bigCI X_{(r^*, c_j)}] = \sum_\mathcal{G} Pr[ I( X_{(r^*, c_i)} ; X_{(r^*, c_j)} ) = 0 | \mathcal{G} ] Pr[ \mathcal{G} ]
$$

If each weighted GPM $\mathcal{G}_k$ is sampled approximately from
some Bayesian posterior $Pr[ \mathcal{G} | O ]$ (and $w_k = 1$
identically), then simple Monte Carlo estimation of the marginal
dependence probability yields an estimate of the {\em posterior}
marginal dependence probability:
$$
Pr[ X_{(r^*, c_i)} \bigCI X_{(r^*, c_j)} | O]
$$

\item $b$ = mutual-information$(\mathcal{G}, {\tt colA} = c_i, {\tt colB} = c_j)$

  The mutual information between two columns can be estimated by the
  standard reduction to KL divergence \citep{cover2012elements}. This
  complements the marginal dependence probability, providing one
  measure of the strength of a dependence.

\end{enumerate}

For convenience, some of the quantities that are ordinarily accessed
via {\tt ESTIMATE} are also made available via {\tt SELECT}.

\subsubsection{{\tt INFER}: summarizing distributions with point estimates.}

Predictive modeling applications sometimes require access to point
predictions rather than samples from predictive distributions. BQL
provides these capabilities using the {\tt INFER} primitive. The
difference between {\tt INFER} and {\tt SELECT} is that {\tt INFER}
incorporates automatic implicit imputation from the underlying
collection of GPMs, plus filtering based on user-specified confidence
thresholds. For simplicity, this paper describes a simplified version
with a single threshold:

\begin{quote}
{\tt INFER {\em target columns} FROM {\em table} [WHERE {\em row filter}] WITH CONFIDENCE {\em confidence level}}
\end{quote}

This operation returns a set of measurements $\{ x^{inf}_{(r,c)} \}$
where unobserved measurements are filled in with point estimates
$\hat{x}_{(r,c)}$ if a prescribed confidence threshold
$p(\mathrm{conf}(X_{(r,c)} = \hat{x}_{(r,c)}) \ge q)$ is reached. More formally:

\begin{flalign*}
x^{inf}_{(r,c)} = \begin{cases}
x_{(r,c)}&\mathrm{foreach}\ (r,c,\cdot) \in O \\
\hat{x}_{(r,c)}&\mathrm{foreach}\ (r,c,\cdot) \notin O\ \mathrm{and}\ p(\mathrm{conf}(X_{(r,c)} = \hat{x}_{(r,c)}) \ge q ) \\
\mathrm{null}&\mathrm{otherwise}
\end{cases}
\end{flalign*}

For discrete measurements, BQL implements $\mathrm{conf}(\cdot)$ in terms of predictive
probability:
$$
\mathrm{conf}(X_{(r,c)} = \hat{x}_{(r,c)}) = p( X_{(r,c)} = \hat{x}_{(r,c)} ) = \sum_{\mathcal{G}} p( X_{(r,c)} = \hat{x}_{(r,c)} | \mathcal{G} ) p( \mathcal{G} ) = \sum_{\mathcal{G}} p( X_{(r,c)} = \hat{x}_{(r,c)} | \mathcal{G} ) w_i
$$
Optimal candidate estimates can be found by optimization, implemented via enumeration:
$$
\hat{x}_{(r,c)} = \argmax_{x} p( X_{(r,c)} = x )
$$

For continuous measurements, there is no canonical definition of
confidence that applies to all GPMs. Here we define $conf(X_{(r,c)} =
x) = q$ as the probability that there is a useful unimodal summary of
the distribution of $X_{(r,c)}$ that captures at least $100q$ percent
of the predictive probability mass. This can be formalized in terms of
mixture modeling. Let $\phi_l$ be the parameters of mixture component
$l$; for continuous data, we will use Gaussian component models, so
$\phi_l = (\mu_l, \sigma_l)$. Let $\pi_l$ be the mass associated with
component $l$. We will choose $conf(\cdot)$ and $\hat{x}$ as
follows:

\begin{flalign*}
\{(\phi_l, \pi_l)\} & \sim p( \{(\phi_l, \pi_l)\} | \{ X_{(r,c)}^k \} )\ \mathrm{for}\ 0 \le k \le K^+ \\
l^* & = \argmax_{l} \pi_l \\
\hat{X}_{(r,c)} & = \mu_{l^*} \\
conf(X_{(r,c)} & = \hat{x}_{(r,c)}) = \pi_{l^*}
\end{flalign*}

Note that this approach can recover the behavior of the chosen
strategy for discrete data by using component models that place all
their probability mass on single values. The current prototype
implementation of BayesDB uses a standard Gibbs sampler for a
Dirichlet process mixture model \citep{crosscatjmlr, neal98,
  rasmussen00} to sample $\{(\phi_l, \pi_l)\} | \{ X_{(r,c)}^k \}$,
with $K^+ = 1000$ by default. Adjusting $K^+$ and the amount of
inference done in this mixture model can yield a broad class of
tradeoffs between time, accuracy, and variance; the current values are
chosen for simplicity.


\subsection{Model and data independence}

Relational databases revolutionized the processing and analysis of
business data by enabling a single centrally managed data base to
shared by multiple applications and also shared between operational
and analytic workloads. This in turn accelerated the development of
high performance and efficient databases, because the common
abstraction became a target for researchers and industrial
practitioners looking to build high performance system software with a
broad impact. The relational model enabled sharing and infrastructure
reuse because interactions with the data, queries, are expressed in a
notation (most popularly SQL) that is independent of the physical
representation of the data \citep{codd1970relational}. Without this
independence, physical data layout must be carefully tailored to
particular workloads, specialized code written to manipulate the
layout, and these data formats and access methods cannot easily be
shared.

BayesDB aims to provide additional abstraction barriers that insulate
clients from the statistical underpinnings of data analysis. Clients
need to be able to specify data analysis steps and workflows in a
notation that is independent of the models and runtime inference
strategies used to implement individual primitives, and (where
possible) the modeling strategies used to produce models from the
original data.

Recall that the complete persistent state of a single population in
BayesDB is characterized by two mathematical objects:

\begin{enumerate}

\item The complete set of observed measurements $O = \{(r_i, c_i,
  x_{(r_i,c_i)})\}$.
\item The weighted collection of GPMs $\{(w_k, \mathcal{G}_k)\}$. Note that this
  notation makes no commitment as to the content of the GPMs, the
  weights, or the procedures by which they were obtained.

\end{enumerate}

The independencies provided by BayesDB can be described in terms of
these objects:

\begin{enumerate}

\item {\em Physical data independence.} The notation for $O$ makes no
  commitment as to the physical representation of the measurements.
  The definitions of BQL primitives given above therefore do not
  depend on details of the data representation to define their
  values. However, as with SQL, small changes in representation may
  yield large changes in runtime performance.

\item {\em Physical model independence.} The notation for each $(w_k,
  \mathcal{G}_k)$ makes no commitment as to the specific probability
  distribution induced over the set of random measurements $X =
  \{X_{(r_i,c_i)}\}$. The definitions of BQL primitives given above
  therefore do not depend on the details of the probabilistic models
  used to define the random result set for each query. In principle,
  the mathematical properties of the models as well as their software
  implementation (or even implementing platform) can be changed
  without invalidating end user queries. However, small changes in the
  generative population model may yield large changes in the results
  of $simulate$ and $logpdf$.

\end{enumerate}

Databases provide other finer-grained independence properties that may
have useful analogs in BayesDB. For example, let us partition the
random variables induced by a given GPM into two subsets $X^A =
\{X_{(r^a_i,c^a_i)}\}$ and $X^B = \{X_{(r^b_j,c^b_j)}\}$. An example
of a desirable data-dependent independence property is that if $X^A|O
\bigCI X^B|O$ in the ``true'' GPM, then $Q | X^A, X^B = Q | X^A$ in
any inferred models. Informally, this rests on the model-building
strategy: if the model-builder recovers the correct independencies,
then the independence of query results follows. This can be thought of
as an analogue of logical data independence, which stipulates that
e.g.\ adding new features should not affect the behavior of existing
applications whose results do not depend on the value of these new
features. Formalizing and verifying these properties is an important
challenge for future research.

%
%
%
%
%
%
%
%

%
%

\section{Modeling with the Meta-Modeling Language}

BayesDB also provides the Meta-Modeling Language (MML), a
probabilistic programming language for building models of data
tables. MML programs consist of {\em modeling tactics} that control
the behavior of an automatic model-building engine. These tactics take
several forms: statistical datatypes; initialization of weighted
collections of random models; approximately Bayesian updating of the
model collection; qualitative assertions about dependence and
independence between variables; and the use of custom statistical
models for specific conditional distributions. All these tactics are
currently implemented in terms of a unifying semi-parametric Bayesian
model that fills in all unspecified aspects.

\subsection{Statistical datatypes}

This metadata constrains the probability models that will be used for
each column of data and can also be used to choose appropriate
visualizations. It is straightforward to support several different
kinds of data:

\begin{enumerate}

\item {\em Categorical values from a closed set.} This datatype
  includes a dictionary that maps the raw data values (often strings)
  to canonical numerical indexes for efficient storage and
  processing. This information can also be used to inform modeling
  tactics. For example, in the current version of MML, closed-set
  categorical variables are modeled generatively via a multinomial
  component model with a symmetric Dirichlet prior on the parameters
  \citep{crosscatjmlr}. Discriminative models for closed-set
  categorical columns could potentially use a multinomial logit link
  function, or an appropriate multi-class classification scheme.

\item {\em Binary data.} Data of this type is generatively modeled
  using an asymmetric Beta-Bernoulli model \citep{crosscatjmlr}
  that can better handle sparse or marginally biased variables than a
  symmetric alternative. Also, a broad class of discriminative
  learning techniques can natively handle the binary classification
  problems induced by binary variables.

\item {\em Count data.} Non-negative counts can be naturally modeled
  generatively by a Poisson-Gamma model or discriminatively by a GLM
  with the appropriate link function.

\item {\em Numerical data.} By default, data of this type is
  generatively modeled using a standard Normal-Gamma model. It is
  straightforward to add numerical ranges to enforce truncation
  post-hoc, and to add numerical pre-transformations that are
  appropriate for data that is naturally viewed as normal only on a
  log scale.

\end{enumerate}

We have performed preliminary experiments on other datatypes built on
standard statistical models. For example, cyclic data can be handled
via a von Mises model \citep{gopal2014mises}. Many other datatypes can
be handled by the appropriate generalized linear model
\citep{mccullagh1989generalized}. Broadening the set of primitive data
types and assessing coverage on a representative corpus of real-world
databases will be a key research challenge going forwards.

\subsection{Bayesian generative population meta-models}

Some data generators can be learned from data. Often the learning
mechanism will be based on approximate probabilistic inference in a
{\em meta-model}: a probabilistic model defined over a space of data
generators, each of which is also a probabilistic model in its own
right. Thus far, all BayesDB meta-models have been {\em Markov chain
  meta-models}. These meta-models internally maintain a single sample
from an approximate posterior, and provide a Markov chain transition
operator that updates this sample stochastically. 

\begin{enumerate}

\item $\mathcal{G} = (\theta_{\mathcal{G}}^0,
    \mathbf{X}_{\mathcal{G}}) =$ initialize(meta-schema = $\Lambda$)

  Initializes a new meta-model with arbitrary parameters and an
  associated tabular data store.

\item incorporate(id = $r$, values = $\{(c_j, x_{(r,c_j)})\}$)

  Creates a new member of the population with the given row index and values and
  stores it. Errors result from duplicate indexes or variables $c_j$
  whose values $x_{(r,c_j)}$ are not compatible with the meta-schema
  $\Lambda$ (e.g. because the expected data type is incompatible with
  a provided value).

\item remove(id = $r$)

  Removes a member of the population from the data store.

\item infer(program = $\mathcal{P}$)

  Simulate an internal Markov chain transition operator $\mathcal{T}$
  to improve the quality of the current sampled model representation:
  $$\theta_{\mathcal{G}}^{i+1} = \mathcal{T}(\theta_{\mathcal{G}}^i)$$

\end{enumerate}

Some Markov chain meta-models are {\em asymptotically Bayesian},
i.e. the distribution that results from sequences of $T$ updates
converges to the posterior over meta-models as $T$ goes to infinity:
$$
\lim_{t \to \infty} \mathrm{D}_{KL}( p( \theta_{\mathcal{G}} | \mathbf{X}_{\mathcal{G}} ) || p( \mathcal{T}^{t}(\theta_{\mathcal{G}}) ) ) = 0
$$

A sufficiently expressive Markov chain meta-model may also be
asymptotically consistent in the usual sense. The default
semi-parametric GPM provided by BayesDB is designed to be both
asymptotically consistent and asymptotically Bayesian; these
invariants are crucial for its robustness, broad applicability, and
suitability for use by non-experts. Formally specifying and validating
these properties is an important challenge for future research.

\subsubsection{Controlling inference via INITIALIZE and ANALYZE.}

The MML allows users to control the process by which models are
created and updated to reflect the data. These capabilities are
exposed via two commands:

\begin{enumerate}

\item {\tt INITIALIZE {\em k} MODELS FOR {\em population}}

  This command creates models by sampling their structure and
  parameters from the underlying GPM's prior. This is implemented by
  delegation to {\tt initialize($\Lambda$)} where $\Lambda$ is the
  entire MML schema so far.

\item {\tt ANALYZE [{\em variable subset} OF] {\em population} FOR
    {\em timelimit}}

  This command performs approximately Bayesian updating of the models
  in the weighted collection by delegating to the {\tt infer()}
  procedure from the underlying GPM. Here is a typical invocation:

\begin{quote}
{\tt ANALYZE my$\_$population FOR 10 MINUTES}
\end{quote}

When no variable subset is provided, analysis is done on all the
latent variables associated with every GPM in the weighted
collection. Finer-grained control is also possible using variable
subset specifiers that pick out particular portions of the latent state in the GPM; these details are beyond the scope of this paper.


\end{enumerate}

\subsection{Qualitative constraints}

The BayesDB MML provides constructs for specifying qualitative
constraints on the dependence and independence relationships
\citep{pearl88}. The model-building engine attempts to enforce them in
all GPMs\footnote{The current implementation does not attempt to
  detect contradictions.}. These constraints are specified as follows:

\begin{quote}
{\tt ALTER METAMODEL FOR {\em population} ENSURE {\em colA} IS
  [NOT] MARGINALLY DEPENDENT ON {\em colB}}
\end{quote}

It is also possible to {\tt INITIALIZE} and {\tt ANALYZE} models that
do not respect the constraints, and then enforce them after the fact:

\begin{quote}
{\tt ALTER MODELS FOR {\em population} ENSURE {\em colA} IS [NOT] MARGINALLY DEPENDENT ON {\em colB}}
\end{quote}

These commands enable domain experts to apply qualitative knowledge to
make better use of sparse data. This can be crucial for improving
analysis and model credibility in the eyes of domain experts. They
also create the opportunity for false or unjustified knowledge to
influence the results of analysis. This can reduce credibility in the
eyes of statisticians or domain skeptics who want to see all
assumptions in an analysis scrutinized empirically.

\subsection{Incorporating foreign statistical models}

A crucial aspect of MML is that it permits experts to override the
automatic model-building machinery using custom-built statistical
models. Feedforward networks of such models can be specified as
follows:

\begin{quote}
{\tt ALTER SCHEMA FOR {\em population} MODEL {\em output variables} GIVEN {\em input variables} USING FOREIGN PREDICTOR FROM {\em source file}}
\end{quote}

Presently these models are presumed to be discriminative. They are
only required to be able to simulate from a probability distribution
over the output variables conditioned on the inputs, and to evaluate
the probability density induced by this distribution.

\subsection{A semi-parametric factorial mixture GPM}

The current implementation of MML implements all the above commands in
terms of approximate inference in single, unusually flexible,
semi-parametric Bayesian meta-model. This GPM is closely related to
CrossCat \citep{crosscatjmlr}. The CrossCat model is a factorial
Dirichlet process mixture model, where variables are assigned to
specific Dirichlet process mixtures by inference in another Dirichlet
process mixture model over the columns. The version used for
implementing MML adds two key components:

\begin{enumerate}

\item {\em Deterministic constraints on model structure.} Users can
  specify constraints on the marginal dependence or independence of
  arbitrary pairs of variables.

\item {\em Feedforward networks of discriminative models conditioned
    on the outputs of the generative model.} This allows users to
  combine general-purpose density estimation with standard statistical
  techniques such as regression as well as complex computational
  models with noisy outputs.

\end{enumerate}

Thus in MML, the CrossCat probability model is used as the root node
in a directed graphical model. Each other node in the graph
corresponds to specific discriminative model, directly conditioned on
the inputs of its immediate ancestors. Undirected terms attached to
the root node enforce deterministic constraints.

It is helpful to view this model in terms of a ``divide and conquer''
modeling strategy that bottoms out in foreign predictors and other
standard parametric models from Bayesian statistics:

\begin{enumerate}

\item All variables not explicitly assigned to a custom model are
  divided into marginally independent groups. Variables in the same
  group are assumed to be marginally dependent. Partitions of
  variables that do not respect the given marginal dependence and
  independence constraints are rejected. Each group of variables
  induces an independent subproblem that will typically be far lower
  dimensional than the original high-dimensional problem.

\item For each subproblem, divide the rows into clusters whose values
  are marginally dependent given any variable-specific
  hyperparameters.

\item For each cluster, use a simple product of parametric models ---
  i.e. a ``naive Bayes'' approach \citep{duda2001pattern} --- to
  estimate the joint distribution.

\end{enumerate}

Inference in the GPM thus addresses modeling tradeoffs that resemble
the decisions faced in exploratory analysis, confirmatory analysis,
and predictive modeling. The most crucial decisions involve defining
which subset of the data is relevant for answering each question. A
secondary issue is what probabilistic model to use for each subset;
absent prior knowledge, these are chosen generically, based on the
type of the data in the column.

\subsubsection{A ``divide-and-conquer'' generative process}


The generative process that induces the default GPM can be described
using the following notation:

\begin{tabular}{c}\\\end{tabular}

\hspace{-0.4in}\begin{tabular}{|l|l|}
\hline
  {\bf Name} & {\bf Description} \\
\hline
  $\alpha_D$ & Concentration hyperparameter for CRP that slices the columns \\
  $\vec{\lambda}_d$ & Hyperparameters for column $d$ (datatype-dependent) \\
  $z_d$ & Slice (column partition) assigned to column $d$ \\
  $\alpha_v$ & Concentration hyperparameter for CRP that clusters rows for slice $v$ \\
  $y^v_r$ & Cluster assigned to row $r$ with respect to slice $v$ \\
  $\vec{\theta_c^d}$ & Model parameters for column $d$ cluster $c$ (datatype-dependent) \\
  $\vec{x}_{(\cdot, d)}^c $ & Values in cluster $c$ for column $d$, i.e. $\{ x_{(r,d)} \mid y|^{z_d}_r = c\}$ \\
  $u_d$ & An indicator such that $u_d = 1$ iff $d$ is modeled by a foreign predictor \\
  $par(d)$ & The set of input dimensions for the foreign predictor \\
  & conditionally modeling variable $d$ \\
  $\vec{\phi_d}$ & Parameters for the foreign predictor conditionally modeling variable $l$ \\
  $m_d(x_{(r,d)}; \vec{\phi_d}, \vec{x}_p)$ & The stochastic model for the foreign predictor used for \\
  & variable $d$ (with density $m^{dens}_d(\cdot)$) with $\vec{x}_p = \{x_{(r,p)} | p \in par(d)\})$\\
  $\delta_m{\vec{z}}$ & Characteristic function enforcing marginal (in)dependence constraint $m$\\
$V_d(\cdot)$ & A generic hyper-prior of the appropriate type for
               variable or dimension $d$.\\

$M_d(\cdot)\ \mathrm{and}\ L_D(\cdot)$& A datatype-appropriate parameter prior (e.g. a Beta prior for
  binary data, \\
$\forall\ d\ s.t.\ u_d =
  1$  & Normal-Gamma for continuous data, or Dirichlet for
  discrete data),
  \\ &  and likelihood model (e.g.\ Bernoulli, Normal or
  Multinomial).\\
\hline
\end{tabular}

\vspace{12pt}

Using this notation, the unconstrained generative process for the
default meta-model can be concisely described in statistician's
notation as follows:
\begin{flalign*}
\alpha_D & \sim \mathrm{Gamma}(k = 1, \theta = 1) &&  \hfill \\
\vec{\lambda}_d & \sim \ V_d(\cdot) & \mathrm{for each}\ d \in \{1, \cdots, D\} \hfill \\
z_d & \sim \ \mathrm{CRP}(\{z_i \mid i \neq d\}; \alpha_D) & \mathrm{for each}\ d \in \{1, \cdots, D\}\\
\alpha_v & \sim \ \mathrm{Gamma}(k = 1, \theta = 1) & \mathrm{for each}\ v \in \vec{z} \\
y^v_r & \sim \ \mathrm{CRP}(\{y^v_i \mid i \neq r\}; \alpha_v) & \mathrm{for each}\ v \in \vec{z}\ \mathrm{and} \\ & & r \in \{1, \cdots, R\} \\
\vec{\theta_c^d} & \sim \ M_d(\cdot; \vec{\lambda}_d) & \hfill \\
\vec{x}_{(\cdot, d)}^c = \{ x_{(r,d)} \mid y^{z_d}_r = c \} & \sim
\prod_r L_d(\vec{\theta_c^d}) & \mathrm{if}\ u_d = 0 \hfill \\
\vec{x}_{(\cdot, d)} = \{ x_{(r,d)} \} & \sim m_d(\vec{\phi_d}; \{x_{(r,p)} | p \in par(d)\}) &
\mathrm{if}\ u_d = 1 \hfill \\
c_m \sim \delta_m{(\vec{z})} && \mathrm{for each\ (in)dependence\ constraint}
\end{flalign*}

The true generative process also must ensure that $c_m = 1$ for all of
the $M$ (in)dependence constraints. This is enforced by conditioning
on the event $\{ c_m = 1\}$. A generative model for this constrained
process can be given trivially by embedding the unconstrained
generative process in the inner loop of a rejection sampler for
$\{c_m\}$ \citep{mansinghka2009natively, murray2009gaussian}.

\subsubsection{The joint density}

Here we use $\theta_{\mathcal{G}}$ to denote all the latent
information in a semi-parametric GPM $\mathcal{G}$ needed to capture
its dependence on the data $O$. This includes the concentration
parameter $\alpha_D$ for the CRP over columns, the variable-specific
hyper-parameters $\{\vec{\lambda}_d\}$, the column partition
$\vec{z}$, the column-partition-specific concentration parameters
$\{\alpha_v\}$ and row partition $\{\vec{y}^v\}$, and the
category-specific parameters $\{\theta_c^d\}$. Note that in this
section, $M_d, V_d, L_d$, and $CRP$ each represent probability density
functions rather than stochastic simulators.

Given this notation, we have: \hspace{-1.0in}%
\begin{flalign*}
P(\theta_{\mathcal{G}}, O) & =
P(\mathbf{X}, \{\vec{\theta_c^d}\}, \{\vec{y}^v, \alpha_v\}, \{\vec{\lambda}_d\}, \vec{z}, \alpha_D) \\
& =  e^{-\alpha_D} \big( \prod_{d \in D} V_d(\vec{\lambda}_d) \big) \mathrm{CRP}(\vec{z} ; \alpha_D) \big( \prod_{v \in \vec{z}} e^{-\alpha_v} \mathrm{CRP}(\vec{y}^v ; \alpha_v) \big) \\
& \times
\big( \prod_{v \in \vec{z}} \prod_{c \in \vec{y}^v} \prod_{d \in \{i \mathrm{\ s.t.\ } z_i = v\}}
M_d(\vec{\theta}_c^d ; \vec{\lambda}_d) \prod_{r \in c} L_d(x_{(r,d)} ; \vec{\theta_c^d}) \big)
\big( \prod_m \delta_m{\vec{z}} \big)
\\ & \times
\big( \prod_{d\ \mathrm{with}\ u_d=1}  \prod_r m^{dens}_d(x_{(r,d)}; \vec{\phi_d}, \{x_{(r,p)} | p \in par(d)) \big)
\end{flalign*}


\subsubsection{Inference via sequential Monte Carlo with Gibbs proposals and Gibbs rejuvenation}

Inference in this meta-model is performed via a sequential Monte Carlo
scheme, in which each row is incorporated incrementally, with all
latent variables proposed from their conditional
distribution. Additionally, clients can control the frequency and
target latent variables for rejuvenation kernels based on Gibbs
sampling, turning the overall scheme into a resample-move algorithm
\citep{andrieu2003introduction, smith2013sequential}. This combination
enables parallel inference and estimation of marginal probabilities
while allowing the bulk of the inferential work to be done via a
suitable Markov chain.

\begin{enumerate}

\item incorporate(id = $r$, values = $\{(c_j, x_{(r,c_j)})\}$)

  Each row is incorporated via a single Gibbs step that numerically
  marginalizes out all the latent variables associated with the row
  \citep{smith2013sequential, murphy2002dynamic}. The associated
  weight is the marginal probability of the measurements to be
  incorporated:

  $$
  w'_i = w_i * p( \{(c_j, x_{(r,c_j)})\} | \mathcal{G_i} )
  $$

  This operation is linear in the number of observed cells for the
  record being incorporated, the number of total slices, and the
  maximum number of clusters associated with any slice.

\item infer(iterations = $N$, type = rows | columns | parameters
    | hyperparameters | foreign | resample, slice = $j$ | NA, cluster
    = $k$ | NA, foreign\_predictor = $l$ | NA)

  This operation applies a particular transition operator, specified
  by the arguments, to a selected subset of the latent variables. Each
  invocation affects all particles in the sequential Monte Carlo
  scheme. By varying the {\tt type} parameter, a client can control
  whether inference is performed over the row-cluster assignment
  variables, the column-slice assignment variables, the cluster
  parameters, the column-specific hyperparameters, or all latent
  variables associated with a specific foreign predictor. An
  invocation with {\tt type = resample} applies multinomial resampling
  to the weighted collection of models.

  This allows for a limited form of inference programming
  \citep{mansinghka2014venture}, as follows. By varying the {\tt
    slice}, {\tt cluster}, or {\tt foreign\_predictor} variables,
  clients can instruct the GPM to only perform inference on a specific
  subset of the latent variables. Computational effort can thus be
  focused on those latent variables that are most relevant for a given
  analysis, rather than uniformly distributed across all latent
  variables in the GPM. This is most useful when the queries of
  interest focus on a subset of the variables, or when the clusters
  are well-separated.


  The prototype implementation of BayesDB uses row-cluster,
  column-slice, cluster-parameter, and column-hyperparameter
  transition operators from \citet{crosscatjmlr}. The only
  modification is that the log joint density now includes terms for
  enforcing each of the (in)dependence constraints, and also terms for
  the likelihood induced by each foreign predictor, as described
  above.

\end{enumerate}

This interface allows clients to specify multiple MCMC, SMC and hybrid
strategies for inference. The default inference program that is
invoked by the {\tt ANALYZE} command in BQL does no resampling and
selects slices and clusters to do inference on via systematic
scans. It thus can be thought of as an MCMC scheme with multiple
parallel chains. This approach is conservative and makes it easier to
assess the stability and reproducibility of inference, although it is
unlikely to be the most efficient approach in some cases.

\section{Discussion}

This paper has described BayesDB, a probabilistic programming platform
that allows users to directly query the probable implications of
statistical data. The query language can solve statistical inference
problems such as detecting predictive relationships between variables,
inferring missing values, simulating probable observations, and
identifying statistically similar database entries. Statisticians and
domain experts can incorporate (in)dependence constraints and custom
models using a qualitative language for probabilistic models. The
default meta-model frees users from needing to know how to choose
modeling approaches, remove records with missing values, detect
outliers, or tune model parameters. The prototype implementation is
suitable for analyzing complex, heterogeneous data tables with up to
tens of thousands of rows and hundreds of variables.

\subsection{Related work in probabilistic programming}

Most probabilistic programming languages are intended for model
specification \citep{goodmanEtal08_church, stan-manual:2015, milch20071,
  pfeffer2009figaro}. This is fundamentally different from BQL and
MML:

\begin{enumerate}

\item In BQL, probabilistic models are never explicitly
  specified. Instead, an implicit set of models is averaged over (or
  sampled from) as needed.

\item With MML, users specify constraints on an algorithm for model
  discovery and need not explicitly select any specific models. These
  constraints generally do not uniquely identify the structure of the
  model that will ultimately be used.

\end{enumerate}

In contrast, with languages such as Stan \citep{stan-manual:2015},
each program corresponds to a specific probabilistic model whose
structure is fixed by the program source. Tabular
\citep{gordon2014tabular}, a probabilistic language designed for
embedding into spreadsheets that applies user-specified factor graph
models defined in terms of observed and latent variables to datasets
represented as sub-tables, seems closest in structure to BQL. However,
like BUGS and Stan, Tabular does not aim to hide the conceptual
vocabulary of probabilistic modeling from its end users, and it
focuses on user-specified models. Other integrations of probabilistic
modeling with databases such as \citep{thoresingh} are similarly
focused on sophisticated modeling but do not provide a
model-independent abstraction for queries or support for general
Bayesian data analysis.

It is straightforward to extend MML to allow syntactic escapes into
all these languages that allow external probabilistic programs to be
used as foreign predictors.

\subsection{Related work in probabilistic databases}

BayesDB takes a complementary approach to several recent projects that
integrate aspects of probabilistic inference with databases. The most
closely related systems are MauveDB \citep{deshpande2006mauvedb} and
BBQ \citep{deshpande2004model}. They provide {\em model-based views}
that enable users to run standard SQL queries on the outputs of
statistical models. These models must be explicitly specified as part
of the schema. This is useful for some machine learning applications
but does not address the core problems of applied inference, such as
data exploration, data cleaning, and confirmatory analysis. Both
systems also use restricted model classes that can easily introduce
substantial for ad-hoc predictive queries.

Other systems such as MLBase \citep{kraska2013mlbase} and GraphLab
\citep{low2012distributed} aim to simplify at-scale development and
deployment of machine learning algorithms.  MLBase and GraphLab host
data in a distributed database environment and provide operators for
scalable ML algorithms. Systems such as SimSQL
\citep{cai2013simulation} and its ancestor, MCDB
\citep{jampani2008mcdb}, provide SQL operators for efficient Monte
Carlo sampling. In principle, several of these systems could serve as
runtime platforms for optimized implementations of BQL and the MML.

\subsubsection{Uncertain data versus uncertain inference}

The database research community has proposed several probabilistic
databases that aim to simplify the management and querying of data
that is ``uncertain'' or ``imprecise'' \citep{dalvi2009probabilistic}.
This ``data uncertainty'' is different from the inferential
uncertainty that motivates BayesDB. Even when the data is known with
certainty, it is rarely possible to uniquely identify a single model
that can be used with complete certainty. Second, each probable model
is likely to have uncertain implications. Extensions of BayesDB that
augment GPMs with probabilistically coherent treatments of data
uncertainty are an important area for future research.

\subsection{Limitations and future work}

Additional GPMs and meta-models are needed for some
applications. There are specialized SQL databases that strike
different tradeoffs between query latency, workload variability, and
storage efficiency. Similarly, we expect that future GPMs and
meta-models will strike different tradeoffs between prediction speed,
prediction accuracy, statistical model capacity, and the amount of
available data. In some cases, the semi-parametric meta-model
presented here may be adequate in principle but producing an
appropriate implementation is a significant systems research
project. For example, it may be possible to build versions suitable
for ad-hoc exploration of distributed databases such as Dremel
\citep{melnik2010dremel} or Spark \citep{zaharia2010spark}. In other
cases, fundamentally different model classes may be more
appropriate. For example, it seems appealing to jointly model
populations of web browsing sessions and web assets with
low-dimensional latent space models \citep{stern2009matchbox}.

It will be challenging to develop query planners that can handle GPMs
given by arbitrary probabilistic programs. A key issue is that the
full GPM interface allows for complex conditional queries over
composite GPMs that may require data-dependent inference
strategies. One potential approach is to specify GPMs as probabilistic
programs in a language with programmable inference; currently, the
only such language is VentureScript. The inference strategy needed to
answer a given query could then be assembled on-demand.

BQL and MML have yet to incorporate key ideas from several significant
subfields of statistics. For example, neither language has explicit
support for causal relationships and arbitrary counterfactuals
\citep{pearl88, pearl2009causality, pearl2001bayesianism}. Both BQL
and MML make the standard, simplistic assumption that data is missing
at random. Neither BQL nor MML has native support for
longitudinal or panel data or for time-series; instead, users must apply
standard workarounds or implement custom data types. A minor
limitation is that hierarchical models are currently supported by
merging subpopulations, retaining an indicator variable, and treating
any variables unique to a given subpopulation as missing. It should
instead be possible to build GPMs that jointly model subpopulations
that are separately represented (and that therefore may not share the
same set of observable variables). It will also be important to
develop a formal semantics and cost model for both BQL and MML.

\noindent {\bf Qualitative probabilistic programming.} BQL and MML are
qualitative languages for quantitative reasoning. They make it
possible for users to perform Bayesian data analysis without needing
to know how to specify quantitative probabilities or model parameters.
However, the set of qualitative constructs that they support is
limited, and needs to be expanded. For example, in MML, it will be
important to support conditional dependence constraints. These
could be specified generatively, e.g. by defining a directed acyclic
graph over subsets of variables, and leaving the model builder to fill
in the (conditional) joint distributions over each subset of
variables. In BQL, it would be interesting to explore the addition of
commands for optimization and decision-theoretic choice, with
objective functions specified both explicitly and implicitly. Finally,
it will be interesting to explore elicitation strategies based on
``programming by example''. For example, users could create datasets
by iteratively specifying prototypical examples and turn them into
large datasets by treating each as the seed for a separate synthetic
population, produced via {\tt SIMULATE}.

\subsection{Conclusion}

Traditional databases protect consumers of data from ``having to know
how the data is organized in the machine'' \cite{codd1970relational} and
provide automated data representations and retrieval algorithms that
perform well enough for a broad class of applications. Although this
abstraction barrier is only imperfectly achieved, it has proved useful
enough to serve as the basis of multiple generations of software and
data systems. This decoupling of task specification from
implementation made it possible to improve performance and reliability
--- of individual database indexes, and in some cases of entire
database systems --- without needing to notify end users. It also
created a simple conceptual vocabulary and query language for data
management and data processing that spread far farther than the
systems programming knowledge needed to implement it.

BayesDB aims to insulate consumers of statistical inference from the
concepts of modeling and statistics and provide a simple, qualitative
interface for solving problems that currently seem quantitative and
complex. It also allows models, analyses, and data resources to be
improved independently. It is not yet clear how deeply the analogy
with traditional databases will run. However, we hope that BayesDB
represents a significant step towards making statistically rigorous
empirical inference more credible, transparent and ubiquitous.

\vskip 0.2in
\bibliography{bib}

\end{document}